\documentclass[runningheads]{llncs}
\usepackage{graphicx}

\usepackage{tikz}
\usepackage{comment}
\usepackage{amsmath,amssymb} 
\usepackage{color}
\usepackage{graphicx}
\usepackage{booktabs}
\usepackage{latexsym}
\usepackage{enumitem}
\usepackage{amsmath}
\usepackage{graphicx}
\usepackage{setspace}
\usepackage{multirow}
\usepackage{pifont}
\usepackage{array}
\usepackage{xcolor,pifont}
\usepackage{tablefootnote}
\usepackage{float}
\usepackage{booktabs}
\usepackage{diagbox}
\usepackage{hyperref}
\usepackage{tikz}
\usepackage{comment}
\usepackage{amsmath,amssymb} 
\usepackage{subcaption}
\usepackage[accsupp]{axessibility}  %
\usepackage[accsupp]{axessibility}  

\newcommand\MyBox[1]{
  \fbox{\lower0.75cm
    \vbox to 1.2cm{\vfil
      \hbox to 1.2cm{\hfil\parbox[c]{1.2cm}{#1}\hfil}
      \vfil}%
  }%
}
\newcommand{\etal}{\textit{et al}.}
\newcommand{\ie}{\textit{i}.\textit{e}.}
\newcommand{\eg}{\textit{e}.\textit{g}.}
\newcommand{\Ie}{\textit{I}.\textit{e}.}
\newcommand{\Eg}{\textit{E}.\textit{g}.}
\begin{document}
\pagestyle{headings}
\mainmatter
\def\ECCVSubNumber{5327}  

\title{A Dataset for Interactive Vision-Language Navigation with Unknown Command Feasibility} 

\titlerunning{Interactive Vision-Language Navigation with Unknown Command Feasibility}
%
\author{Andrea Burns\inst{1} \and
Deniz Arsan\inst{2} \and Sanjna Agrawal\inst{1} \and Ranjitha Kumar\inst{2} \and Kate Saenko\inst{1,3} \and Bryan A. Plummer\inst{1}}
\authorrunning{Burns et al.}
%
\institute{Boston University, Boston MA 02215, USA  \email{\{aburns4,sanjna,saenko,bplum\}@bu.edu} \\
\and
University of Illinois Urbana-Champaign, Champaign IL 61820, USA \\
\email{\{darsan2,ranjitha\}@illinois.edu}\\
\and
MIT-IBM Watson AI Lab, Cambridge MA 02142, USA\\
}
\maketitle
\addtocontents{toc}{\protect\setcounter{tocdepth}{-10}}

\begin{abstract}
Vision-language navigation (VLN), in which an agent follows language instruction in a visual environment, has been studied under the premise that the input command is fully feasible in the environment. Yet in practice, a request may not be possible due to language ambiguity or environment changes. To study VLN with unknown command feasibility, we introduce a new dataset Mobile app Tasks with Iterative Feedback (MoTIF), where the goal is to complete a natural language command in a mobile app. Mobile apps provide a scalable domain to study real downstream uses of VLN methods. Moreover, mobile app commands provide instruction for interactive navigation, as they result in action sequences with state changes via clicking, typing, or swiping. MoTIF is the first to include feasibility annotations, containing both binary feasibility labels and fine-grained labels for why tasks are unsatisfiable. We further collect follow-up questions for ambiguous queries to enable research on task uncertainty resolution. Equipped with our dataset, we propose the new problem of feasibility prediction, in which a natural language instruction and multimodal app environment are used to predict command feasibility. MoTIF provides a more realistic app dataset as it contains many diverse environments, high-level goals, and longer action sequences than prior work.
We evaluate interactive VLN methods using MoTIF, quantify the generalization ability of current approaches to new app environments, and measure the effect of task feasibility on navigation performance.

\end{abstract}

\section{Introduction}
\begin{figure}[t]
\centering
    \includegraphics[scale=0.1615]{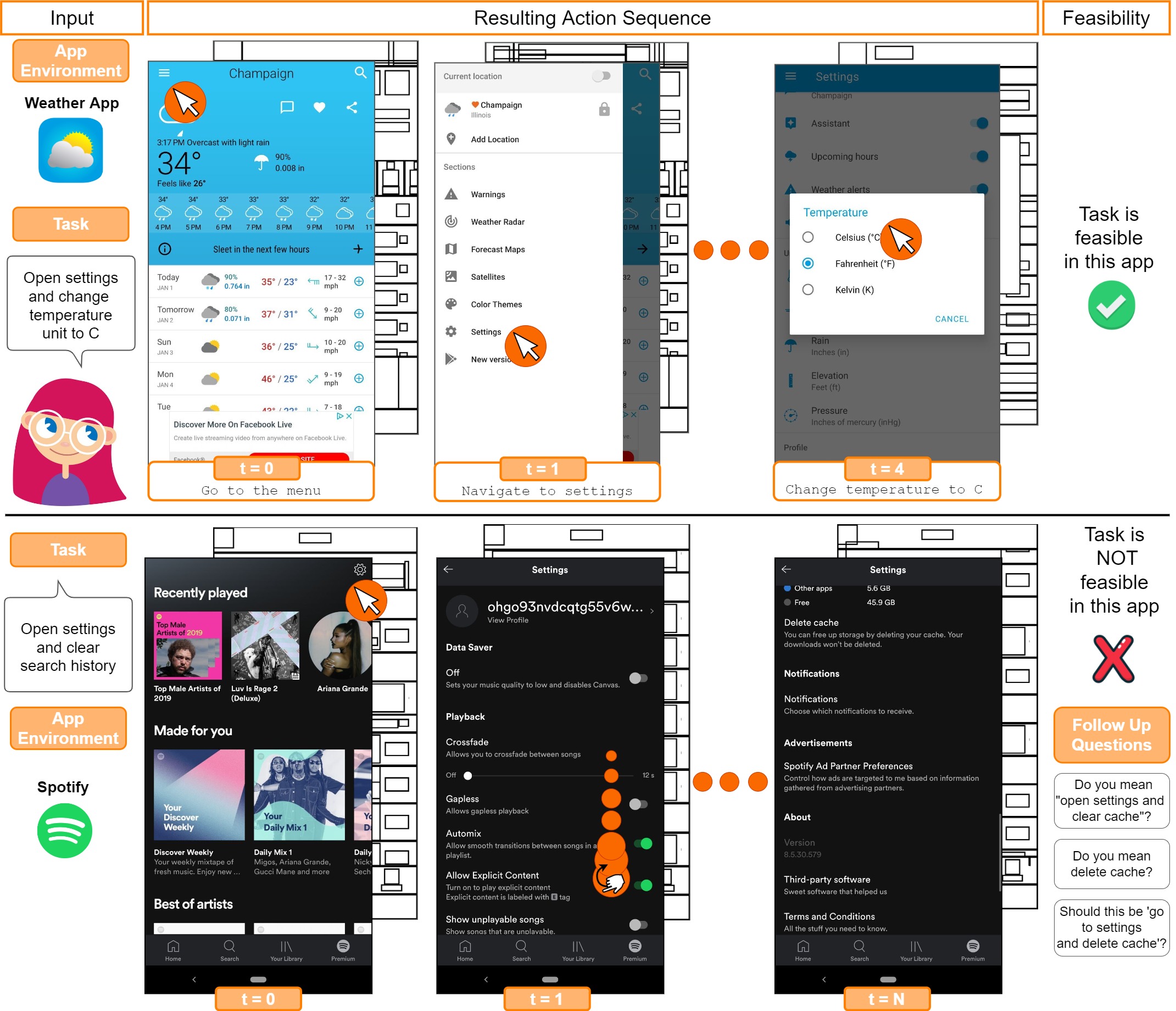}
    \caption{MoTIF natural language commands which may not be possible. At each time step, action coordinates (\ie, where clicking, typing, or scrolling occurs), the app screen, and view hierarchy (\ie, the app backend, illustrated behind it) are captured}
    \label{fig:motivation}
\end{figure}

Vision-language navigation (VLN) has made notable progress toward natural language instruction following~\cite{blukis2021persistent,irshad2021hierarchical,min2021film,nguyen2019hanna,ALFRED20,singh2020moca,vln_ssa}. While navigation datasets exist for home environments~\cite{vln,embodiedqa,ku-etal-2020-room,ALFRED20} and digital environments like mobile apps and websites~\cite{li-etal-2020-mapping,vut,langtoelem,wob}, none capture the possibility that the language request may not be feasible in the given environment. When high-level natural language goals are requested, they may not be feasible for various reasons: the request may be ambiguous or state dependent, refer to functionality that is no longer available, or is reasonable in a similar environment but not satisfiable in the current. Task feasibility has been studied to determine question relevance for text-only~\cite{Gardner2020DeterminingQP} and visual question answering~\cite{vizwiz,massiceti,ray2016question}, but it has not been explored in interactive multimodal environments.

To study interactive task feasibility, we propose Mobile app Tasks with Iterative Feedback (MoTIF)\footnote{\url{https://github.com/aburns4/MoTIF}}, the largest dataset designed to support interactive methods for completing natural language tasks in mobile apps. As illustrated in Figure~\ref{fig:motivation}, a sample includes the natural language command (\ie, task), app view hierarchy, app screen image, and action coordinates for each time step. MoTIF contains both feasible and infeasible requests, unlike any VLN dataset to date. 
In addition to these binary feasibility labels for each task, we collect subclass annotations for why tasks are infeasible and natural language follow-up questions. 
Our dataset provides a domain with practical downstream applications to study vision-language navigation, as well as data for investigating app design~\cite{rico,erica,designsemantics}, human-computer interfaces~\cite{sugilite,convobreak,demoplusLi2021}, and document understanding~\cite{appalaraju2021docformer,Li_2021_CVPR,canvasvae}. 

We propose a baseline model for task feasibility prediction and confirm app exploration is necessary, with visual inputs key to accuracy. Surprisingly, prior representation learning approaches specific to the mobile app domain (\eg, app icon features) do not result in the best performance. 
We then evaluate methods for automating MoTIF's commands and find MoTIF's diverse test set are challenging for prior work. Performance trends between seen and unseen app environments point to the need for more in-app exploration during training and qualitative failures in the best baseline model demonstrate the importance of visual understanding for MoTIF.

We summarize our contributions below:
\begin{itemize}
    \item A new vision-language navigation dataset, Mobile app Tasks with Iterative Feedback (MoTIF). MoTIF has free form natural language commands for interactive goals in mobile apps, a subset of which are infeasible. It contains natural language tasks for the most app environments to date. MoTIF also captures multiple interactions including clicking, swiping and typing actions.
    \item A new vision-language task: interactive task feasibility classification, along with subclass annotations on why tasks are infeasible and follow-up questions for research toward resolving task uncertainty via dialogue. 
    \item Benchmarks for feasibility classification and task automation with MoTIF. A thorough feature exploration is performed to evaluate the role of vision and language in task feasibility. We compare several methods on mobile app task automation, analyze generalization, and examine the effects of feasibility.
\end{itemize}

\section{Related Work}
We now discuss the key differences between MoTIF and existing datasets; we provide a side-by-side comparison in Table~\ref{tab:compare}.
\smallskip 

\noindent\textbf{Task Feasibility}
Vision-language research has recently begun to study task feasibility. Gurari \etal\hspace{0.25mm} introduced VizWiz~\cite{vizwiz}, a visual question answering dataset for images taken by people that are blind, resulting in questions which may not be answerable. To the best of our knowledge, VizWiz is the only vision-language dataset with annotations for task feasibility, but it only addresses question answering over static images. Additionally, images that cannot be used to answer visual questions are easily classified, as they often contain blurred or random scenes (\eg, the floor). 
Gardner \etal ~\cite{Gardner2020DeterminingQP} explored question-answer plausibility prediction, but the questions used were generated from a bot, which could result in extraneous questions also easy to classify as implausible. Both are significantly different from the nuanced tasks of MoTIF with human generated queries, for which exploration is necessary to determine feasibility. MoTIF's infeasible tasks are always relevant to the Android app category, making it more challenging to discern feasibility compared to the distinct visual failures present in VizWiz. 
\smallskip
\begin{table}[t]
  \renewcommand\arraystretch{0.95}
    \centering
        \caption{Comparison of MoTIF to existing datasets. We consider the number of natural language commands, command granularity, existence of feasibility annotations, the number of environments and whether the visual state is included in annotations}
    \begin{tabular}{|l|c|c|c|c|c|c|}
    \hline
       \multirow{3}{*}{Dataset} & \multicolumn{3}{c|}{Language Annotations} & \multicolumn{2}{c|}{Dataset Environment} \\ 
    \cline{2-6}
         & \# Human & Task & \multirow{2}{*}{Feasibility} & \# & Visual \\ 
                  & Annotations & Granularity & & Environments & State \\ 
                                              \hline

                            \textbf{(a) House} & & & & & \\
                          \hspace{1mm} R2R~\cite{vln} & 21,567 & Low & \color{red}{\ding{55}} & 90 &\color{green}{\ding{51}} \\ 
        \hspace{1mm} IQA~\cite{IQA} & \color{red}{\ding{55}} & High & \color{red}{\ding{55}} & 30 &\color{green}{\ding{51}} \\ 
         \hspace{1mm} ALFRED~\cite{ALFRED20} & 25,743 & High \& Low &  \color{red}{\ding{55}}& 120 & \color{green}{\ding{51}} \\ 
         \hline
          \textbf{(b) Webpage} & & & & & \\
       \hspace{1mm} MiniWoB~\cite{wob} &  \color{red}{\ding{55}}  & High & \color{red}{\ding{55}} & 100  & \color{red}{\ding{55}} \\
        \hspace{1mm} PhraseNode~\cite{langtoelem} & 50,000 & Low & \color{red}{\ding{55}} & 1,800 & \color{red}{\ding{55}} \\ 
         \hline
         \textbf{(c) Mobile App} & & & & & \\ 
         \hspace{1mm} RicoSCA~\cite{li-etal-2020-mapping} & \color{red}{\ding{55}} & Low & \color{red}{\ding{55}} & 9,700 &  \color{red}{\ding{55}} \\ 
        \hspace{1mm} PIXELHELP~\cite{li-etal-2020-mapping} & 187 &  Low & \color{red}{\ding{55}} & 4 & \color{red}{\ding{55}}\\ 
        \hspace{1mm} MoTIF (Ours) & 6,100 & High \& Low & \color{green}{\ding{51}} & 125  & \color{green}{\ding{51}} \\ 
         \hline
    \end{tabular}
    \label{tab:compare}
\end{table}

\noindent\textbf{Vision-Language Navigation}
There are datasets that strictly navigate to locations like Room-to-Room~\cite{vln} and Room-Across-Room~\cite{ku-etal-2020-room}, as well as interactive datasets where agents perform actions in the environment to complete a goal like ALFRED~\cite{ALFRED20}. MoTIF is most similar to interactive VLN, as the natural language instructions are intended to complete a goal for the user, which requires clicking, typing, or swiping actions in the environment. However, an advantage of MoTIF is that it is a real, non-simulated domain to study interactive navigation, unlike all VLN prior work which uses simulated data~\cite{IQA,virtualhome8578984,ALFRED20,Zhu2017VisualSP}.
\smallskip

\noindent\textbf{Digital Task Automation}
Prior work has not studied web task automation in a multimodal setting, ignoring the rendered website image~\cite{langtoelem,wob}. The existing datasets MiniWoB~\cite{wob} and PhraseNode~\cite{langtoelem} also lack realism, as MiniWoB consists of handcrafted HTML and PhraseNode only captures single action commands on the home screen of websites. Unlike these datasets which limit interaction to a single screen, MoTIF contains action sequences with many different states (as shown in Figure~\ref{fig:motivation}), with a median of eight visited screens. 

RicoSCA and PIXELHELP were introduced for mobile app task automation by Li \etal~\cite{li-etal-2020-mapping}. RicoSCA makes use of the mobile app dataset Rico~\cite{rico}, which captures random exploration in Android apps. Li \etal\hspace{0.25mm} synthetically generate random commands with templates like ``\textit{click on} \textbf{x}'' and stitch multiple together to any prescribed length. 
These generated step-by-step instructions do not reflect downstream use, where users ask for a high-level goal. For MoTIF, we instead collect free form high-level goals, and then post-process our data to automatically generate the low level subgoal instructions.
PIXELHELP is a small mobile app dataset, but most commands are device specific. \Ie, the tasks refer to the phone itself, such as ``\textit{in the top control menu click the battery saver},'' and are not in-app tasks like those in Figure~\ref{fig:motivation}. 
PIXELHELP also only contains clicking, while MoTIF has clicking, typing and swiping actions.

\section{MoTIF Dataset}
\begin{figure}[t]
    \centering
    \includegraphics[scale=0.17275]{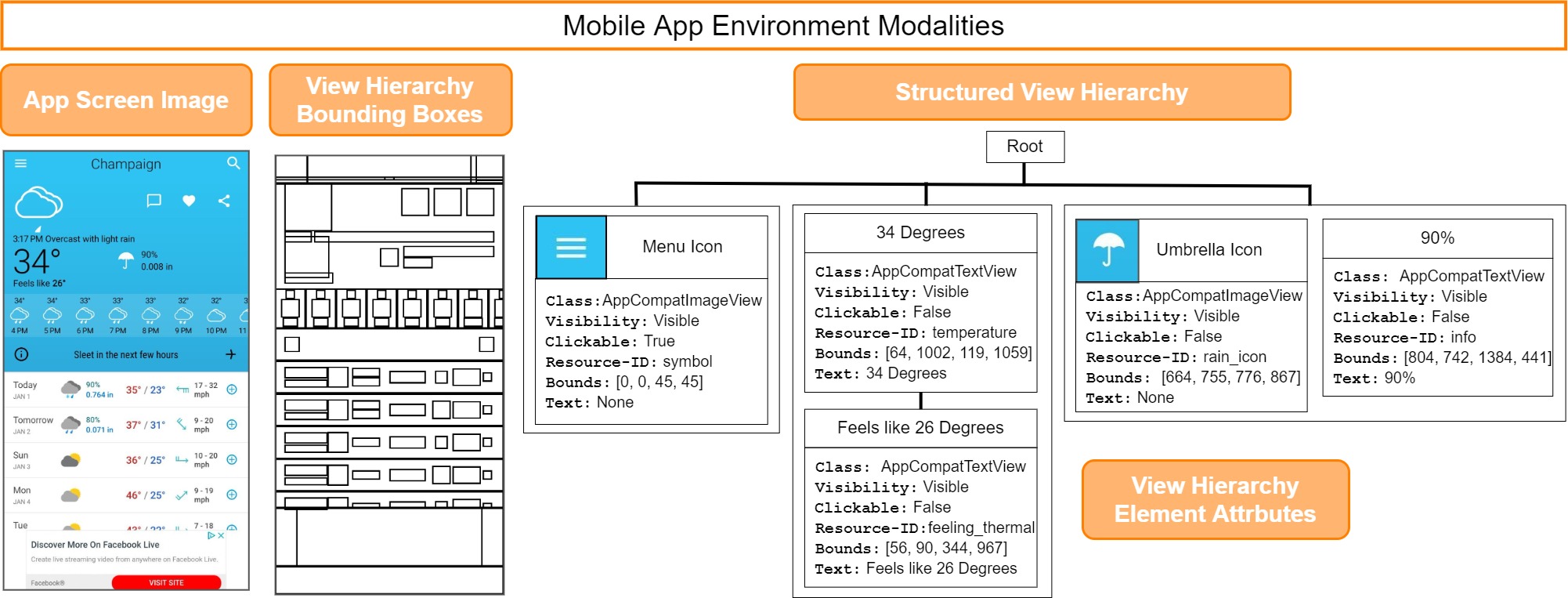}
    \caption{We illustrate captured app modalities: the rendered screen and view hierarchy, which contains element metadata such as the Android class, resource ID, and text}
    \label{fig:modes}
\end{figure}

For a mobile app task dataset, we need natural language tasks for apps and their resulting action sequence. Figure~\ref{fig:motivation} illustrates MoTIF tasks like ``\textit{open settings and change temperature unit to C}.'' For each command, we collect expert demonstrations of attempts to complete the request. At each time step we capture the app screen, the app backend view hierarchy, what type of action is taken, and where the action occurred. We show the modalities captured at each time step in greater detail in Figure~\ref{fig:modes}. The Android app backend, \ie, view hierarchy, is a tree-like structure akin to the Document Object Model (DOM) used for HTML. It organizes each screen element hierarchically, and contains additional metadata like the Android class of an element (\eg, a text view or image view), its resource identifier, the text it contains, whether it is clickable, and other attributes.

\subsection{Data Collection}
\label{sec:data}

\begin{figure}[t]
    \centering
    {\includegraphics[scale=0.2]{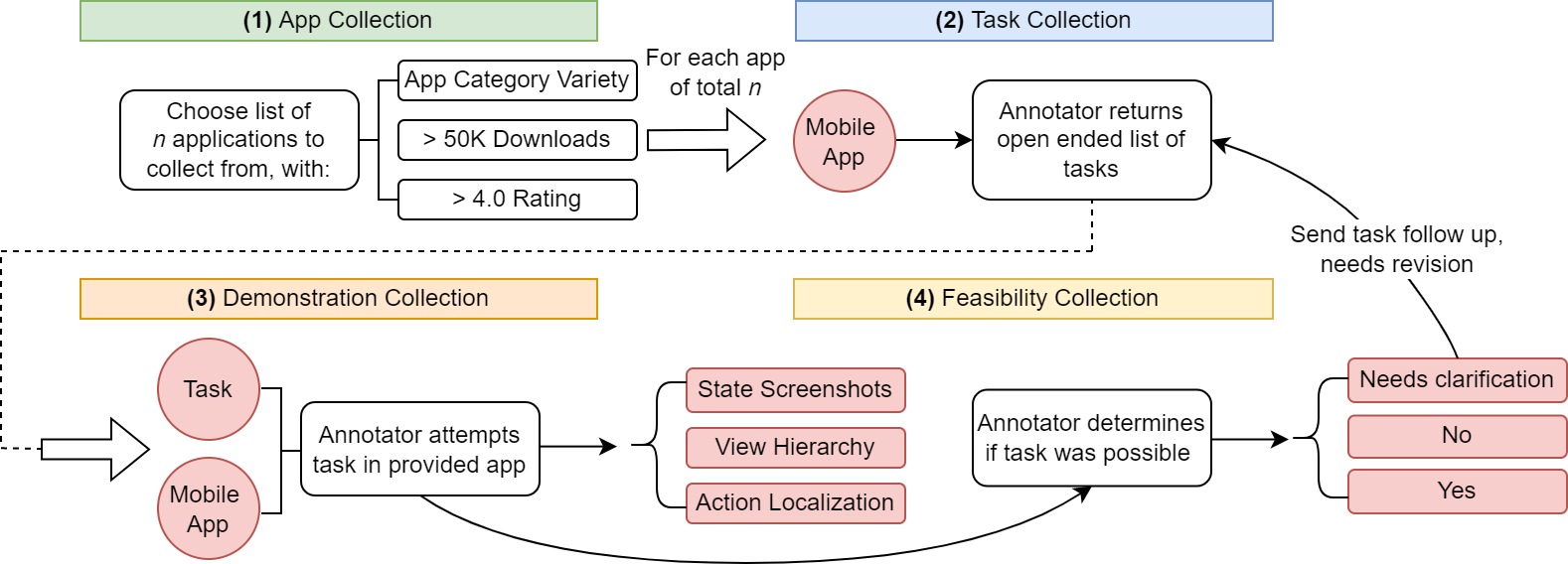}
        \caption{The data collection pipeline (see Section~\ref{sec:data}). Colored boxes (app, task, demonstration, and feasibility collection) are stages of curating the dataset}
    \label{fig:collection}
    }
\end{figure}

We provide a general framework for others to collect natural language data with unknown feasibility; Figure~\ref{fig:collection} illustrates the collection pipeline. We select 125 apps for MoTIF over 15 app categories (the complete app list can be found in the Supplementary). 
Ten apps with (1) at least 50k downloads and (2) a rating higher than or equal to 4/5 were chosen for each category.
Next, a first set of annotators writes commands. A list of (app, task) pairs are then provided to a second set of annotators in an interactive session,
where they attempt the task, specify if it is not feasible, and can ask a clarifying question if not. 
The Supplementary includes annotator demographics, payment, and collection interface details. 


\smallskip
 
\noindent\textbf{Natural Language Commands}
\label{nlanns}
To collect natural language tasks, we instruct workers to write commands as if they are asking the app to perform the task for them.
Annotators can explore the app before deciding on their list of tasks. We ask them to write functional or navigational tasks, and not commands requiring text comprehension like summarizing an article.
We neither structure the written tasks nor prescribe a specific number of tasks to be written for each app.
\smallskip

\noindent\textbf{Task-Application Pairing}
\label{pair}
When collecting natural language tasks, annotators can first explore the app. Once we have tasks for every app, we introduce additional feasibility uncertainty for the demonstration stage by collecting demos for both the original (app, task) list, as well as tasks paired with apps they were not originally written for.  
We create these additional (app, task) pairs by clustering tasks within each Android category (for example, clustering all tasks for Music and Audio Android apps) and selecting representatives from each cluster. These representative tasks are then collected for all apps of that category, which we coin ``\textit{category-clustered}.'' Specifically, we cluster the mean FastText embedding~\cite{conneau2017word} of the language commands using K-Means~\cite{kmeans}.

Clusters are visualized with T-SNE~\cite{vanDerMaaten2008} (see Supplementary). If a particular app's tasks are isolated from other clusters, we retain ``\textit{app-specific}'' pairings, \ie, the (app, task) pairs for tasks specifically written for the given app. This resulted in 40 apps having only app-specific tasks. If two apps' tasks are closely clustered, we group them; 
 17 apps' tasks were gathered this way. 
Figure~\ref{fig:motivation} (bottom) shows a category-clustered task which was deemed infeasible by annotators. The command ``\textit{open settings and clear search history}'' was paired with the music app Spotify even though it was not written for it. This is a sensible request given that Spotify is a music streaming app. Yet, no search history is found under settings, only the option to ``delete cache,'' and follow-up questions are asked.

\label{sec:analysis}



\smallskip

\noindent\textbf{Task Demonstration and Feasibility Annotations}
\label{demofeasanns}
Once the language commands are paired with apps, we instruct new annotators to demonstrate the task in the given app. We provide a website interface connected to physical Android phones for crowd workers to interact with, as well as anonymized login credentials so that no personally identifiable information is collected. They are instructed to record their demonstration after they have logged in (we consider logging in to be a separate task). After attempting to complete the task, they are brought to a post-survey where they provide details on whether or not the task was successfully completed. We therefore have demonstrations of actions taken both in successful and unsuccessful episodes, which may provide interesting insight toward how to reason about whether a task is or is not feasible, and why.

\subsection{Dataset Analysis}
\smallskip

\noindent\textbf{Natural Language Commands}
We collected over 6.1k natural language tasks across 125 Android apps. 
The vocabulary size was 3,763 after removing non-alphanumeric characters. 
The average number of tasks submitted per app is 56, with average length being 5.6 words. The minimum task length is one, consisting of single action tasks like `refresh' or `login,' with the longest at 44 words. 
Word cloud visualizations, additional examples and statistics are in the Supplementary. 
\begin{table}[t]
    \centering
      \renewcommand\arraystretch{0.95}
        \caption{Task feasibility and follow-up question breakdown. Annotators can state the action: can't be completed (impossible), is under-specified (unclear), may be possible, but are unsure how or other tasks need to be completed first (premature)}
    \begin{tabular}{|c|c|c|c|c|c|}
    \hline
      \multirow{2}{*}{\#} & \multirow{2}{*}{Feasible}  &  \multicolumn{3}{c|}{Infeasible} & \multirow{2}{*}{Total} \\
      \cline{3-5}
      & & Impossible & Unclear & Premature & \\
      \hline
     Task Demonstrations & 3,337 & 911 & 159 & 300 & 4,707 \\ 
      \hline
      Follow-Up Questions & 93 & 253 & 136 & 164 & 646 \\ 
      \hline
    \end{tabular}
    \label{tab:possible}
\end{table}
\smallskip

\noindent\textbf{Feasibility Annotations}
We collect at least five expert demonstrations per (app, task) pair for two purposes: to reach a majority feasibility label and to capture different attempts of the same task, as some tasks can be completed in multiple ways. See the Supplementary for an annotator agreement histogram.

Of the resulting tasks, 29.2\% are deemed infeasible by at least five crowd workers. 
However, the tasks considered infeasible do not always correlate to mismatched (app, task) pairs, \ie, some 
\textit{app-specific} tasks are deemed infeasible during demonstration. This confirms the need to study commands with unknown feasibility, as someone familiar with an app can still pose requests that are either not possible, ambiguous, or state dependent. Of the infeasible tasks, 16.8\% are from app-specific pairs. 
\Eg, the request ``\textit{click shuttle and station}'' originally written for the NASA app was labeled infeasible because the app has changing interactive features. Thus app changes and dynamic features also motivate studying infeasible requests, as a task that was once feasible may not always be.

Table~\ref{tab:possible} provides statistics on the number of task demonstrations and follow-up questions per feasibility category. There are three options for annotators to choose from: (1) the action cannot be completed in the app, (2) the action is unclear or under-specified, or (3) the task seems to be possible, but they cannot figure out how or other tasks need to be completed first. These map to Table~\ref{tab:possible}'s impossible, unclear, and premature columns. 
If a crowd worker cannot complete the task, they are prompted to ask a follow-up question. We instruct them to write the question(s) such that if they had the answer, they may now be able to complete the original action or perform an alternative task for the user. 



\section{Task Feasibility Experiments}
We first perform experiments with MoTIF for task feasibility. 
Given a natural language command and the app states visited during its demonstration, the purpose of task feasibility prediction is to classify if the command can be completed. To determine feasibility, we expect a model to learn the most relevant state for the requested task and if the functionality needed to complete it is present. 
Our results provide an initial upper bound on performance, as the input action sequences can be considered the ground truth exploration needed to determine feasibility, as opposed to a learned agent's exploration.  
MoTIF has 4.7k demonstrations and we reserve 10\% for testing. Note that our test set only includes (app, task) pairs for which all annotators agreed on their feasibility annotation.

\subsection{Models} We propose a Multi-Layer Perceptron (MLP) baseline with two hidden layers that outputs a binary feasibility prediction. Each MLP is trained for 50 epochs with cross entropy using Stochastic Gradient Descent with a learning rate of 1e-2. The natural language command is always input to the classifier, and we ablate which app environment features are additional input. In addition to the feature ablations, we ablate how the demonstration sequence is aggregated (averaging or concatenating over time steps or using the last hidden state of an LSTM~\cite{lstm}). 
\smallskip

\noindent\textbf{Features}
We encode the task command and view hierarchy elements per step with mean pooled features. Specifically, we try both FastText~\cite{bojanowski-etal-2017-enriching} and CLIP~\cite{clip} (trained with a Transformer backbone for its image and text encoders~\cite{dosovitskiy2021image,transformer}). 
As seen in Figure~\ref{fig:modes}, the view hierarchy captures all rendered app elements and their attributes: the element's text (ET), resource-identifier (ID) and class labels (CLS) which provide content and type information. We use the best combination of these attributes in Table~\ref{tab:feas} and have more ablations in the Supplementary. We also include Screen2Vec~\cite{screen2vec} in our view hierarchy representations. Screen2Vec is a semantic embedding of the view hierarchy, representing the view hierarchy with a GUI, text, and layout embedder. The GUI and text encoders make use of BERT features while the layout features are learned with an autoencoder. Thus, it tries to encode both textual and structural features, but no visual information.

For visual features, we extract ResNet152~\cite{resnet} features for ten crops of each app image and CLIP features of each whole app image. We also include icon features by cropping all icon images per screen (\eg, the menu and umbrella icons shown in Figure~\ref{fig:modes}). We embed each icon image using the embedding layer of a CNN trained for the downstream task of icon classification by Liu \etal~\cite{designsemantics}.
\smallskip

\noindent\textbf{Metrics}
We report the average F1 score over ten runs with different random initialization. ``Infeasible'' is defined as the positive class, as we care more about correctly classifying tasks that are infeasible, than misclassifying feasible tasks.

\subsection{Results}
\begin{table}[t]
        \caption{Task feasibility F1 score using our MLP. We ablate input features and action sequence aggregation. The random baseline predicts a feasibility label given the train distribution. On the right is a confusion matrix for the predictions of our best classifier}
\begin{minipage}{0.6\textwidth}
    \centering
          \renewcommand\arraystretch{0.95}

    \begin{tabular}{|l|c|c|c|}
    \hline
       \multirow{2}{*}{\textbf{C}$_{feas}$ Input Features} & \multicolumn{3}{c|}{Demo Aggregation}\\
       \cline{2-4}
       & Avg & Cat & LSTM \\
       \hline
       \textbf{Random} & \multicolumn{3}{c|}{20.1} \\
       \hline
     \textbf{(a) View Hierarchy} & & & \\ 
     FastText~\cite{bojanowski-etal-2017-enriching} (ET, ID) & 16.7 & 43.6 & 34.1 \\ 
     CLIP~\cite{clip} (ET, ID) & 28.0 & 50.9 & 36.2 \\
     Screen2Vec~\cite{screen2vec} & 25.9 & 33.7 & 36.0 \\ 
      \hline
      \textbf{(b) App Screen Image} & & & \\
     ResNet~\cite{resnet} & 31.3 & 41.9 & 35.9 \\ 
    Icons~\cite{designsemantics} & 0.4 & 40.0 & 15.2 \\
    CLIP~\cite{clip} & 44.7 & 58.2 & \underline{42.8} \\
      \hline
      \textbf{(c) Best Combination} & & & \\
      CLIP~\cite{clip} (Screen, ET, ID) & \underline{44.8} & \underline{61.1} & 40.9 \\
      \hline
    \end{tabular}
\end{minipage}
\begin{minipage}{0.4\textwidth}
\centering

\begin{tabular}{c >{}r @{\hspace{0.7em}}c @{\hspace{0.4em}}c @{\hspace{0.7em}}l}
\hline
\multicolumn{4}{c}{\multirow{2}{*}{\textbf{{(c) Best Combination}}}} \\  \\
\hline
\\
  \multirow{12}{*}{{\rotatebox{90}{\textbf{Prediction}}}} & 
    & \multicolumn{2}{c}{\textbf{Ground Truth}} & \\
  & & Feasible & Infeasible \\
  & \rotatebox{90}{\hspace{-7.5mm}Feasible} & \MyBox{\hspace{2.25mm}76.4\%} & \MyBox{\hspace{3mm}8.6\%} \\[2.4em]
  & \rotatebox{90}{\hspace{-8.5mm}Infeasible} & \MyBox{\hspace{2.5mm}4.0\%} & \MyBox{\hspace{2.5mm}11.0\%} \\
\end{tabular}
\end{minipage}
\label{tab:feas}
\end{table} 

\label{feasperf}
Our best task feasibility classifier (Table~\ref{tab:feas}(c) left) achieves an F1 score of 61.1 when CLIP embeds the task, view hierarchy, and app screen image. This is still fairly low, and feature ablations demonstrate room to improve both the language and visual representations. While CLIP has shown significant performance gains in other vision-language tasks, it is somewhat surprising that domain-specific embeddings (\eg, Screen2Vec, Icons) are not as competitive. The combination of view hierarchy and app screen features does not largely outperform the app screen image CLIP results (and does worse with LSTM aggregation), suggesting a need for better vision-language encodings which can pull features together from different modalities such as the view hierarchy. 

We include the confusion matrix on the right of Table~\ref{tab:feas} for our best model. In downstream use, the classifier would result in 5\% of tasks being missed out on; \ie, 5\% of tasks were incorrectly classified as infeasible. This reduces the utility of assistive applications, where we'd like all possible commands to correctly be completed. However, the 44\% of infeasible tasks that were incorrectly classified as feasible can have more negative consequences. In application, this means a vision-language model would attempt to complete an unsatisfiable request, resulting in unknown behavior. We need downstream models to behave in reliable ways, especially for users that cannot verify the task was reasonably completed.

Table~\ref{tab:feas}(a) left compares methods of encoding the view hierarchy. Using CLIP for view hierarchy elements results in notably better performance than FastText, albeit less significant when input demos are aggregated with an LSTM. Our final view hierarchy embedding is Screen2Vec which performs worse than CLIP and on par with FastText, despite being trained on mobile app data. Screen2Vec may not capture enough low level view hierarchy information to predict feasibility, and methods trained on huge data, even if from another domain, are more powerful.

In Table~\ref{tab:feas}(b) left we ablate over the visual representations of the app screen.
While icon representations are trained on images from the same domain as MoTIF, they are significantly less effective than ResNet and CLIP. The F1 score nearly drops to zero when the average icon feature is used, illustrating that the average icon does not carry useful information for feasibility classification. Icon features may be too low-level or require improved aggregation methods.

Comparing demonstration aggregation methods (averaging, concatenating, or LSTM), there is a trend that concatenating time steps is the best method, suggesting a sequential representation of the action sequence is needed.
However, when the best representations for the view hierarchy and app screen are combined in Table~\ref{tab:feas}(c), averaging manages to outperform the LSTM performance. 

In future work we hope to learn hierarchical representations in order to encode global information such as that of Screen2Vec as well as local information from icon embeddings. Taking advantage of the tree structure from the view hierarchy via Transformers or Graph Neural Networks may help learn structured app features. Additionally, all current approaches do not take into account any notion of app ``affordance,'' \ie, which app elements are most actionable.
\section{Task Automation Experiments}
In app task automation, we are given an app environment (with all of its modalities) and a language command. The goal is to interact with the app and output a sequence of app actions that complete the task, akin to interactive VLN. At each time step there are two predictions: an action (clicking, typing, or swiping) and a localization (grounding visually on the app screen or classifying over the app elements). We benchmark several methods and analyze performance below. 
\subsection{Models} 
We adapt three models for the mobile app domain with as few changes as possible. The VLN approaches described below (Seq2Seq and MOCA) take both the high-level goal and low level instructions as input while Seq2Act only supports low level instruction. In the supplementary we include input language ablations to consider what performance with real downstream use would look like.
\smallskip

\noindent\textbf{Seq2Seq} is a VLN model for the ALFRED, a dataset of actionable commands for tasks in household environments. 
It originally predicts an action and binary mask at each time step. The mask isolates the household object on which the action is performed. The features at each time step include the attended language instruction, the current step's visual features, the last predicted action, and the hidden state of a BiLSTM which takes the former as input. The previous step's BiLSTM hidden state attends to the language input. 
These features are passed to a fully connected layer with Softmax for action prediction and a deconvolutional network for mask prediction. 
We replace the mask prediction network for three fully connected layers that predict a point in the app screen and minimize the mean squared error. Action prediction is trained via cross entropy.
\smallskip

\noindent\textbf{MOCA}~\cite{singh2020moca}, also proposed for ALFRED, decouples the action and grounding predictions of each step in a VLN sequence. One model stream is for the action prediction policy, and another for interaction grounding. Both streams first use a BiLSTM language encoder, which take the high-level goal or low level instruction as input, respectively. The encoded tokens are attended to using ResNet visual features via dynamic attention filters. Then, two LSTM decoders are used: one for the action policy stream and another for the interaction grounding.

At test time MOCA makes use of an off-the-shelf object segmentation model to perform grounding given the predicted object class. To adapt the object class prediction to mobile apps, we instead perform app element type prediction (prediction is over twelve classes, including button, checkbox, edit text, image view, and more). As no such segmentation model exists for mobile apps yet, we also predict bounding box localization directly using the LSTM decoder output, but use the app element type prediction to narrow grounding options at evaluation. 

\smallskip

\noindent\textbf{Seq2Act}~\cite{li-etal-2020-mapping} 
models mobile app task automation in two stages: action phrase extraction and action grounding. Both stages are modeled with Transformers. 
The first model predicts a span (\ie, substring) of the original input command that corresponds to the action type, action location, and action input. 
It has an encoder-decoder architecture: the encoder embeds the instruction's text tokens and the decoder computes a query vector for the action type, location, and input phrases 
given the previously decoded spans. A text span is selected for each decoder query (action type, action location, action input) via cosine similarity. 


The action grounding model takes each extracted phrase as input to predict an action type and location (which app element it is performed on). Actions are predicted given the encoder embedding of the predicted action type span via a Multi-Layer Perception. To localize the action, a Transformer is trained to embed app elements using the view hierarchy attributes as shown in Figure~\ref{fig:modes}.
A Softmax is applied to the similarities of the predicted app location span embeddings and the latent app element representations. 
The max scoring app element becomes the grounding prediction for that time step.
\smallskip

\noindent\textbf{Datasets} We evaluate task automation on two MoTIF test splits: an app seen and an app unseen split to study generalization to new environments; generalization of tasks across apps is provided in the Supplementary. We jointly train VLN models on MoTIF and RicoSCA for additional data (see the Supplementary for additional experiments trained solely on MoTIF). Seq2Act was originally trained on RicoSCA and we adapt its training data split to be able to evaluate seen versus unseen apps at test time.
\smallskip

\noindent\textbf{Features} Visual features for Seq2Seq and MOCA are from the last convolutional layer of a ResNet18, as done for the original models; these features are needed for meaningful localization on the mobile app screen.
We also include CLIP features of the screen at each time step. Note that VLN methods require a test-time environment; we build an offline version of each Android app to approximate a complete state-action space graph. Details on the creation of these graphs can be found in the Supplementary. Seq2Act does not use off-the-shelf features as input; all text and app element embeddings are learned from scratch. 

\begin{table}[t]
    \centering
          \renewcommand\arraystretch{0.95}

        \caption{Mobile app task accuracy on MoTIF. We evaluate the Seq2Seq and MOCA navigation models and the Transformer grounding model Seq2Act}
    \begin{tabular}{|l|c|c|c|c|c|c|}
    \hline
    \multirow{3}{*}{Model} & \multicolumn{3}{c|}{App Seen} & \multicolumn{3}{c|}{App Unseen}\\
    \cline{2-7}
   & \multirow{2}{*}{Action} & \multirow{2}{*}{Ground} & Action + &  \multirow{2}{*}{Action} & \multirow{2}{*}{Ground} & Action +\\
   & & & Ground& & & Ground\\
    \hline
    \textbf{(a) Seq2Seq}~\cite{ALFRED20} & & & & & &  \\
    \hspace{5mm} Complete Sequence & 68.5 & 22.5 & 22.5 & 54.3 & 18.0 & 17.7 \\
    \hspace{5mm} Partial Sequence & 89.5 & 40.4 & 40.1 & 81.7 & 31.3 & 30.6 \\
    \hline
    \textbf{(b) MOCA}~\cite{singh2020moca} & & & & & &\\ 
    \hspace{5mm} Complete Sequence & 51.1 & 21.3 & 20.7 & 44.8 & 17.0 & 15.1 \\
    \hspace{5mm} Partial Sequence & 78.5 & 40.0 & 38.6 & 72.2 & 32.7 & 30.0 \\
    \hline
    \textbf{(c) Seq2Act}~\cite{li-etal-2020-mapping} & & & & & &\\
    \hspace{5mm} Complete Sequence & \underline{97.3} & \underline{32.4} & \underline{32.4} & \underline{96.8} & \underline{28.3} & \underline{28.3} \\
   \hspace{5mm} Partial Sequence  & \underline{99.2} & \underline{66.4} & \underline{66.3} & \underline{99.6} & \underline{67.7} & \underline{67.6}  \\
    \hline
    \end{tabular}
    \label{tab:automate_full}
\end{table}
\smallskip

\noindent\textbf{Metrics}
We report complete and partial sequence accuracy per~\cite{li-etal-2020-mapping}: the complete score for an action sequence is 1 if the predicted and ground truth sequences have the same length and the same predictions at each step, else 0. The partial sequence score is the fraction of predicted steps that match the ground-truth. These are reported for action prediction (Action), action grounding (Ground), or both jointly. 
Seq2Seq localization is correct if the predicted point falls within the bounding box of the ground truth app element. MOCA localization is correct if the predicted bounding box and ground truth have an IoU greater than 0.5.
\subsection{Results}
Despite MOCA being a more recent model for interactive vision-language navigation, it generally does not outperform Seq2Seq.
The app element type prediction MOCA uses may be responsible for the similar or lower accuracy, as the original intention of object class prediction was to narrow down grounding interaction to very few options. \Eg, in the home environments of ALFRED, which MOCA evaluated on, the object class predicted may be apple. If there is only a single apple in the scene, the object segmentation model would be highly effective for grounding. The mobile app domain differs in that there are many app elements per time step of the same type, \eg, there are many app icons or app buttons, and this prediction may not significantly reduce the grounding prediction space.


The Seq2Seq and MOCA models perform worse than Seq2Act. While additional model ablations may improve performance, it is clear that action localization on the continuous app screen is more challenging.
Seq2Act achieves the highest performance for all metrics. Seq2Act was originally evaluated with PIXELHELP~\cite{li-etal-2020-mapping} and achieved 70.6\% complete Action + Ground accuracy on it, much higher than the accuracy reported on MoTIF. This may be due to PIXELHELP containing click-only tasks for four test environments, which does not reflect the model's performance on a greater variety of apps or tasks. MoTIF's step-by-step instructions also contain location descriptions for app elements which don't contain text, differing from the Seq2Act training data distribution.

\begin{figure}[t]
    \centering
    \includegraphics[scale=0.15]{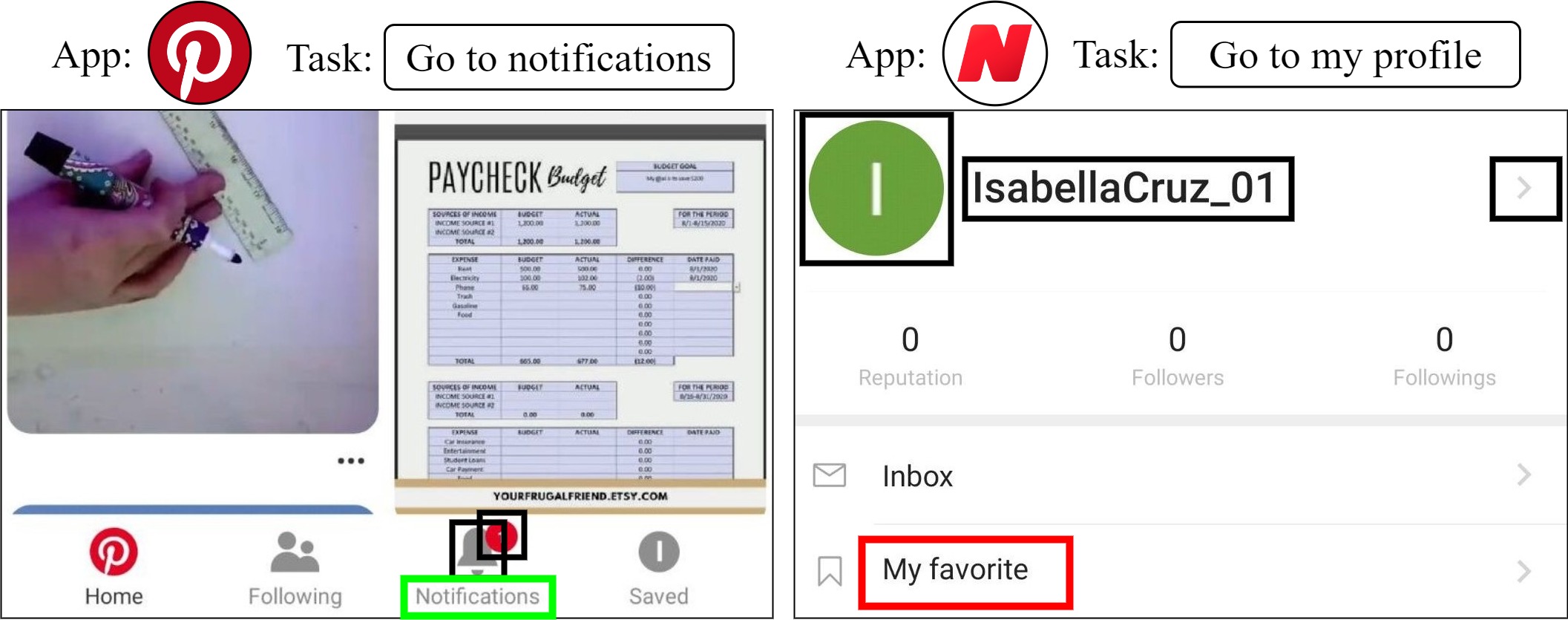}
    \caption{Seq2Act text matching. Green and red boxes are valid and invalid predictions, respectively; black are additional valid ground truth. The left shows valid text matching, identifying ``notifications'' in the app Pinterest. The right shows Seq2Act incorrectly matching ``my'' in the input task to the app element ``My favorite'' in the Opera news app}
    \label{fig:qual}
\end{figure}

Qualitatively inspecting misclassifications, we find one culprit to be Seq2Act overly relying on matching input task text to view hierarchy text. In Figure~\ref{fig:qual}, we show Seq2Act's text matching tendency, which can result in failure. For example, Seq2Act predicts the app element with the word ``my'' in it for the input command ``\textit{go to my profile}.'' These results, in addition to the high visual performance from the feasibility classifier, verifies the need for visual input to correct model bias to match input task text directly to the view hierarchy.  

Performance is unsurprisingly worse for unseen app environments. 
We suspect that current model formulations do not learn enough about app elements outside of the ground truth action sequences during training. None of the benchmark models include exploration, and as a result, may be biased to the small subset of elements seen in expert demonstration. In future work, using pretrained generic app features or incorporating exploration into the training process through reinforcement learning approaches may alleviate this.

\section{Discussion}
We find our best task feasibility prediction results to be low at a 61.1 F1 score, given that the input demonstrations serve as the oracle exploration needed to determine feasibility. In addition to improving vision-language feasibility reasoning, a necessary next step is to instead use learned explorations during training. Our ablations demonstrate that visual inputs are useful for feasibility prediction, and research toward better mobile app features that actually use the rendered screen could increase performance. 
Building hierarchical visual and textual features may provide stronger context clues for determining command feasibility in the app environment. We also hope to perform experiments classifying why tasks are not feasible and automating question generation in response, making use of MoTIF's subclass annotations for infeasible tasks and follow up questions.

By evaluating action and grounding performance independently, we found that models for completing mobile app tasks can have more difficulty grounding and consistently perform more poorly in new app environments. 
Better representations of app elements are needed;
specifically, incorporating pretraining tasks for improved app features or allowing for exploration outside of ground truth action sequences may be necessary to diversify test time predictions.

\smallskip
\noindent\textbf{Limitations}
The MoTIF dataset is not on the scale of pretraining datasets used in other VL tasks (\eg, Alt-Text~\cite{jia2021scaling}, JFT-300M~\cite{jft300}), as it is very expensive and time costly to collect natural language commands and feasibility labels. MoTIF is nonetheless useful to the research community as it can be used to evaluate how existing methods would solve language-based digital tasks in realistic settings.

\smallskip
\noindent\textbf{Societal Impact}
Methods for automating language commands and predicting command feasibility can be used to assist people who are not able to interact with apps, either situationally (\eg, while driving) or physically (\eg, users who are low-vision or blind). Improving mobile app task automation could better balance the capabilities of current assistive technologies, which typically lack agency or flexibility~\cite{screenreader}. \Eg, screen readers are primarily used for web browsing and information consumption (lacking agency), while interactive virtual assistants (\eg, Siri, Alexa) have limited, structured commands (lacking flexibility). 

MoTIF's collection was designed to ensure no personally identifiable information is captured. But, in downstream use of app task automation, user privacy is of concern. People who use assistive tools (\eg, people who are blind) already expose sensitive information to other humans to receive help~\cite{visprivacy,Akter2020IAU}. To mitigate potential harm, deployment of our research can be limited to apps which do not require log in information; these are less likely to include name, address, or payment data. MoTIF does not have tasks which require payment, and we can deny payment related tasks to prevent fraud and other undesired outcomes. 

\section{Conclusion}
\label{sec:conc}
We introduced Mobile app Tasks with Iterative Feedback (MoTIF), a new VLN dataset that contains natural language commands for tasks in mobile apps which may not be feasible. MoTIF is the first dataset to capture task uncertainty for interactive visual environments and contains greater linguistic and visual diversity than prior work, allowing for more research toward robust vision-language methods. We introduced the task of feasibility prediction and evaluate prior methods for automating mobile app tasks. Results verify that MoTIF poses new vision-language challenges, and that the vision-language community can make use of more realistic data to evaluate and improve upon current methods.

\smallskip
\noindent\textbf{Acknowledgements} This work is funded in part by Boston University, the Google Ph.D. Fellowship program, the MIT-IBM Watson AI Lab, the Google Faculty Research Award and NSF Grant IIS-1750563.

%
%
\bibliographystyle{splncs04}
\bibliography{egbib}

\begin{thebibliography}{10}
\providecommand{\url}[1]{\texttt{#1}}
\providecommand{\urlprefix}{URL }
\providecommand{\doi}[1]{https://doi.org/#1}

\bibitem{visprivacy}
Ahmed, T., Hoyle, R., Connelly, K., Crandall, D., Kapadia, A.: Privacy concerns
  and behaviors of people with visual impairments. In: Proceedings of the 33rd
  Annual ACM Conference on Human Factors in Computing Systems. p. 3523–3532.
  CHI '15, Association for Computing Machinery, New York, NY, USA (2015).
  \doi{10.1145/2702123.2702334}, \url{https://doi.org/10.1145/2702123.2702334}

\bibitem{Akter2020IAU}
Akter, T., Dosono, B., Ahmed, T., Kapadia, A., Semaan, B.C.: ``{I} am
  uncomfortable sharing what {I} can't see'': Privacy concerns of the visually
  impaired with camera based assistive applications. In: USENIX Security
  Symposium (2020)

\bibitem{vln}
Anderson, P., Wu, Q., Teney, D., Bruce, J., Johnson, M., S{\"{u}}nderhauf, N.,
  Reid, I.D., Gould, S., van~den Hengel, A.: Vision-and-language navigation:
  Interpreting visually-grounded navigation instructions in real environments.
  In: Proceedings of the IEEE Conference on Computer Vision and Pattern
  Recognition (CVPR) (2018)

\bibitem{appalaraju2021docformer}
Appalaraju, S., Jasani, B., Kota, B.U., Xie, Y., Manmatha, R.: Docformer:
  End-to-end transformer for document understanding. In: 2021 IEEE/CVF
  International Conference on Computer Vision (ICCV) (2021)

\bibitem{blukis2021persistent}
Blukis, V., Paxton, C., Fox, D., Garg, A., Artzi, Y.: A persistent spatial
  semantic representation for high-level natural language instruction execution
  (2021)

\bibitem{bojanowski-etal-2017-enriching}
Bojanowski, P., Grave, E., Joulin, A., Mikolov, T.: Enriching word vectors with
  subword information. Transactions of the Association for Computational
  Linguistics  \textbf{5} (2017)

\bibitem{conneau2017word}
Conneau, A., Lample, G., Ranzato, M., Denoyer, L., J{\'e}gou, H.: Word
  translation without parallel data. In: International Conference on Learning
  Representations (ICLR) (2018)

\bibitem{embodiedqa}
Das, A., Datta, S., Gkioxari, G., Lee, S., Parikh, D., Batra, D.: {E}mbodied
  {Q}uestion {A}nswering. In: Proceedings of the IEEE Conference on Computer
  Vision and Pattern Recognition (CVPR) (2018)

\bibitem{rico}
Deka, B., Huang, Z., Franzen, C., Hibschman, J., Afergan, D., Li, Y., Nichols,
  J., Kumar, R.: Rico: A mobile app dataset for building data-driven design
  applications. In: 30th Annual Symposium on User Interface Software and
  Technology (UIST) (2017)

\bibitem{erica}
Deka, B., Huang, Z., Kumar, R.: Erica: Interaction mining mobile apps. In: 29th
  Annual Symposium on User Interface Software and Technology (UIST) (2016)

\bibitem{dosovitskiy2021image}
Dosovitskiy, A., Beyer, L., Kolesnikov, A., Weissenborn, D., Zhai, X.,
  Unterthiner, T., Dehghani, M., Minderer, M., Heigold, G., Gelly, S.,
  Uszkoreit, J., Houlsby, N.: An image is worth 16x16 words: Transformers for
  image recognition at scale (2021)

\bibitem{Gardner2020DeterminingQP}
Gardner, R., Varma, M., Zhu, C., Krishna, R.: Determining question-answer
  plausibility in crowdsourced datasets using multi-task learning. In:
  W-NUT@EMNLP (2020)

\bibitem{IQA}
Gordon, D., Kembhavi, A., Rastegari, M., Redmon, J., Fox, D., Farhadi, A.:
  {IQA}: Visual question answering in interactive environments. In: 2018
  IEEE/CVF Conference on Computer Vision and Pattern Recognition (CVPR). pp.
  4089--4098 (2018). \doi{10.1109/CVPR.2018.00430}

\bibitem{vizwiz}
Gurari, D., Li, Q., Stangl, A.J., Guo, A., Lin, C., Grauman, K., Luo, J.,
  Bigham, J.P.: Vizwiz grand challenge: Answering visual questions from blind
  people. In: Conference on Computer Vision and Pattern Recognition (CVPR)
  (2018)

\bibitem{resnet}
He, K., Zhang, X., Ren, S., Sun, J.: Deep residual learning for image
  recognition. In: 2016 IEEE Conference on Computer Vision and Pattern
  Recognition (CVPR). pp. 770--778 (2016). \doi{10.1109/CVPR.2016.90}

\bibitem{lstm}
Hochreiter, S., Schmidhuber, J.: Long short-term memory. Neural Comput.
  \textbf{9}(8),  1735–1780 (Nov 1997). \doi{10.1162/neco.1997.9.8.1735},
  \url{https://doi.org/10.1162/neco.1997.9.8.1735}

\bibitem{irshad2021hierarchical}
Irshad, M.Z., Ma, C.Y., Kira, Z.: Hierarchical cross-modal agent for robotics
  vision-and-language navigation. In: Proceedings of the IEEE International
  Conference on Robotics and Automation (ICRA) (2021),
  \url{https://arxiv.org/abs/2104.10674}

\bibitem{jia2021scaling}
Jia, C., Yang, Y., Xia, Y., Chen, Y.T., Parekh, Z., Pham, H., Le, Q.V., Sung,
  Y., Li, Z., Duerig, T.: Scaling up visual and vision-language representation
  learning with noisy text supervision (2021)

\bibitem{ku-etal-2020-room}
Ku, A., Anderson, P., Patel, R., Ie, E., Baldridge, J.: Room-across-room:
  Multilingual vision-and-language navigation with dense spatiotemporal
  grounding. In: Proceedings of the 2020 Conference on Empirical Methods in
  Natural Language Processing (EMNLP). pp. 4392--4412. Association for
  Computational Linguistics, Online (Nov 2020).
  \doi{10.18653/v1/2020.emnlp-main.356},
  \url{https://aclanthology.org/2020.emnlp-main.356}

\bibitem{Li_2021_CVPR}
Li, P., Gu, J., Kuen, J., Morariu, V.I., Zhao, H., Jain, R., Manjunatha, V.,
  Liu, H.: Selfdoc: Self-supervised document representation learning. In: 2021
  IEEE/CVF Conference on Computer Vision and Pattern Recognition (CVPR) (2021)

\bibitem{sugilite}
Li, T.J.J., Azaria, A., Myers, B.A.: Sugilite: Creating multimodal smartphone
  automation by demonstration. In: Proceedings of the 2017 CHI Conference on
  Human Factors in Computing Systems. p. 6038–6049. CHI '17, Association for
  Computing Machinery, New York, NY, USA (2017)

\bibitem{convobreak}
Li, T.J.J., Chen, J., Xia, H., Mitchell, T.M., Myers, B.A.: Multi-Modal Repairs
  of Conversational Breakdowns in Task-Oriented Dialogs, p. 1094–1107.
  Association for Computing Machinery, New York, NY, USA (2020)

\bibitem{demoplusLi2021}
Li, T.J.J., Mitchell, T.M., Myers, B.A.: Demonstration + Natural Language:
  Multimodal Interfaces for GUI-Based Interactive Task Learning Agents, pp.
  495--537. Springer International Publishing, Cham (2021)

\bibitem{screen2vec}
Li, T.J.J., Popowski, L., Mitchell, T.M., Myers, B.A.: Screen2vec: Semantic
  embedding of gui screens and gui components. In: Proceedings of the SIGCHI
  Conference on Human Factors in Computing Systems. CHI '21 (2021)

\bibitem{li-etal-2020-mapping}
Li, Y., He, J., Zhou, X., Zhang, Y., Baldridge, J.: Mapping natural language
  instructions to mobile {UI} action sequences. In: Proceedings of the 58th
  Annual Meeting of the Association for Computational Linguistics. pp.
  8198--8210. Association for Computational Linguistics, Online (Jul 2020).
  \doi{10.18653/v1/2020.acl-main.729},
  \url{https://www.aclweb.org/anthology/2020.acl-main.729}

\bibitem{vut}
Li, Y., Li, G., Zhou, X., Dehghani, M., Gritsenko, A.A.: {VUT:} versatile {UI}
  transformer for multi-modal multi-task user interface modeling. CoRR
  \textbf{abs/2112.05692} (2021), \url{https://arxiv.org/abs/2112.05692}

\bibitem{designsemantics}
Liu, T.F., Craft, M., Situ, J., Yumer, E., Mech, R., Kumar, R.: Learning design
  semantics for mobile apps. In: 31st Annual Symposium on User Interface
  Software and Technology (UIST) (2018)

\bibitem{kmeans}
Lloyd, S.: Least squares quantization in pcm. In: IEEE Transactions on
  Information Theory (1982)

\bibitem{vanDerMaaten2008}
van~der Maaten, L., Hinton, G.: Visualizing data using {t-SNE}. Journal of
  Machine Learning Research  \textbf{9},  2579--2605 (2008),
  \url{http://www.jmlr.org/papers/v9/vandermaaten08a.html}

\bibitem{massiceti}
Massiceti, D., Dokania, P.K., Siddharth, N., Torr, P.H.S.: Visual dialogue
  without vision or dialogue. CoRR  \textbf{abs/1812.06417} (2018),
  \url{http://arxiv.org/abs/1812.06417}

\bibitem{min2021film}
Min, S.Y., Chaplot, D.S., Ravikumar, P., Bisk, Y., Salakhutdinov, R.: Film:
  Following instructions in language with modular methods (2021)

\bibitem{nguyen2019hanna}
Nguyen, K., Daum{\'e}~III, H.: Help, anna! visual navigation with natural
  multimodal assistance via retrospective curiosity-encouraging imitation
  learning. In: Proceedings of the Conference on Empirical Methods in Natural
  Language Processing (EMNLP) (November 2019)

\bibitem{langtoelem}
Pasupat, P., Jiang, T.S., Liu, E.Z., Guu, K., Liang, P.: Mapping natural
  language commands to web elements. In: Empirical Methods in Natural Language
  Processing (EMNLP) (2018)

\bibitem{virtualhome8578984}
Puig, X., Ra, K., Boben, M., Li, J., Wang, T., Fidler, S., Torralba, A.:
  Virtualhome: Simulating household activities via programs. In: 2018 IEEE/CVF
  Conference on Computer Vision and Pattern Recognition (CVPR). pp. 8494--8502.
  IEEE Computer Society, Los Alamitos, CA, USA (jun 2018).
  \doi{10.1109/CVPR.2018.00886},
  \url{https://doi.ieeecomputersociety.org/10.1109/CVPR.2018.00886}

\bibitem{clip}
Radford, A., Kim, J.W., Hallacy, C., Ramesh, A., Goh, G., Agarwal, S., Sastry,
  G., Askell, A., Mishkin, P., Clark, J., Krueger, G., Sutskever, I.: Learning
  transferable visual models from natural language supervision. CoRR
  \textbf{abs/2103.00020} (2021), \url{https://arxiv.org/abs/2103.00020}

\bibitem{ray2016question}
Ray, A., Christie, G., Bansal, M., Batra, D., Parikh, D.: Question relevance in
  vqa: Identifying non-visual and false-premise questions (2016)

\bibitem{wob}
Shi, T., Karpathy, A., Fan, L., Hernandez, J., Liang, P.: World of bits: An
  open-domain platform for web-based agents. In: 34th International Conference
  on Machine Learning (ICML) (2015)

\bibitem{ALFRED20}
Shridhar, M., Thomason, J., Gordon, D., Bisk, Y., Han, W., Mottaghi, R.,
  Zettlemoyer, L., Fox, D.: {ALFRED: A Benchmark for Interpreting Grounded
  Instructions for Everyday Tasks}. In: The IEEE Conference on Computer Vision
  and Pattern Recognition (CVPR) (2020), \url{https://arxiv.org/abs/1912.01734}

\bibitem{singh2020moca}
Singh, K.P., Bhambri, S., Kim, B., Mottaghi, R., Choi, J.: Factorizing
  perception and policy for interactive instruction following. In: Proceedings
  of the IEEE/CVF International Conference on Computer Vision (ICCV) (2021)

\bibitem{jft300}
Sun, C., Shrivastava, A., Singh, S., Gupta, A.: Revisiting unreasonable
  effectiveness of data in deep learning era. CoRR  \textbf{abs/1707.02968}
  (2017), \url{http://arxiv.org/abs/1707.02968}

\bibitem{transformer}
Vaswani, A., Shazeer, N., Parmar, N., Uszkoreit, J., Jones, L., Gomez, A.N.,
  Kaiser, L., Polosukhin, I.: Attention is all you need. In: Conference on
  Neural Information Processing Systems (NeurIPS) (2017)

\bibitem{screenreader}
Vtyurina, A., Fourney, A., Morris, M.R., Findlater, L., White, R.W.: Bridging
  screen readers and voice assistants for enhanced eyes-free web search. In:
  International ACM SIGACCESS Conference on Computers and Accessibility
  (ASSETS) (2019)

\bibitem{canvasvae}
Yamaguchi, K.: Canvasvae: Learning to generate vector graphic documents. In:
  Proceedings of the IEEE/CVF International Conference on Computer Vision
  (ICCV) (2021)

\bibitem{vln_ssa}
Zhu, F., Zhu, Y., Chang, X., Liang, X.: Vision-language navigation with
  self-supervised auxiliary reasoning tasks. In: 2020 IEEE/CVF Conference on
  Computer Vision and Pattern Recognition (CVPR). pp. 10009--10019 (2020).
  \doi{10.1109/CVPR42600.2020.01003}

\bibitem{Zhu2017VisualSP}
Zhu, Y., Gordon, D., Kolve, E., Fox, D., Fei-Fei, L., Gupta, A.K., Mottaghi,
  R., Farhadi, A.: Visual semantic planning using deep successor
  representations. In: 2017 IEEE International Conference on Computer Vision
  (ICCV) (2017)

\end{thebibliography}
\section*{\Large Supplementary}


\section{MoTIF Collection}
\label{sec:collection}

For data collection, we use UpWork\footnote{\url{https://www.upwork.com/}} as our crowd sourcing platform and hired 34 people to collect our dataset. Of the annotators, 21 identified as female and 13 identified as male. The median age of the annotators was 23.5 years old. Annotators were from 18 different states in the U.S. and had a range of education from a high school diploma to a master's degree (2 have high school degrees, 24 have bachelor's degrees, and 8 have master's degrees).

Annotators were selected on UpWork if their profile skills listed data entry. As the initial iteration of MoTIF is in English, we also required annotators be fluent in English, but did not require them to be native speakers. We posted separate job listings for the task writing (base rate \$15/hr) and task demonstration (base rate \$10/hr) portions of the data collection, having independent annotators for the two stages. Annotators hired for the task writing portion were not informed of our interest in potentially ambiguous or infeasible tasks.

For the annotators hired for task demonstration, we additionally required them to have personal experience with Android devices so that there was no additional noise introduced from people unfamiliar with Android apps. We created anonymized login information for annotators so that no personally identifiable information was collected. Additional interface details and an example of the interface used by the workers (Figure~\ref{fig:website}) is provided in Section~\ref{interface}. 
\subsection{Data Collection Interface}
\label{interface}
\begin{figure*}[t]
    \centering
    \includegraphics[scale=0.3]{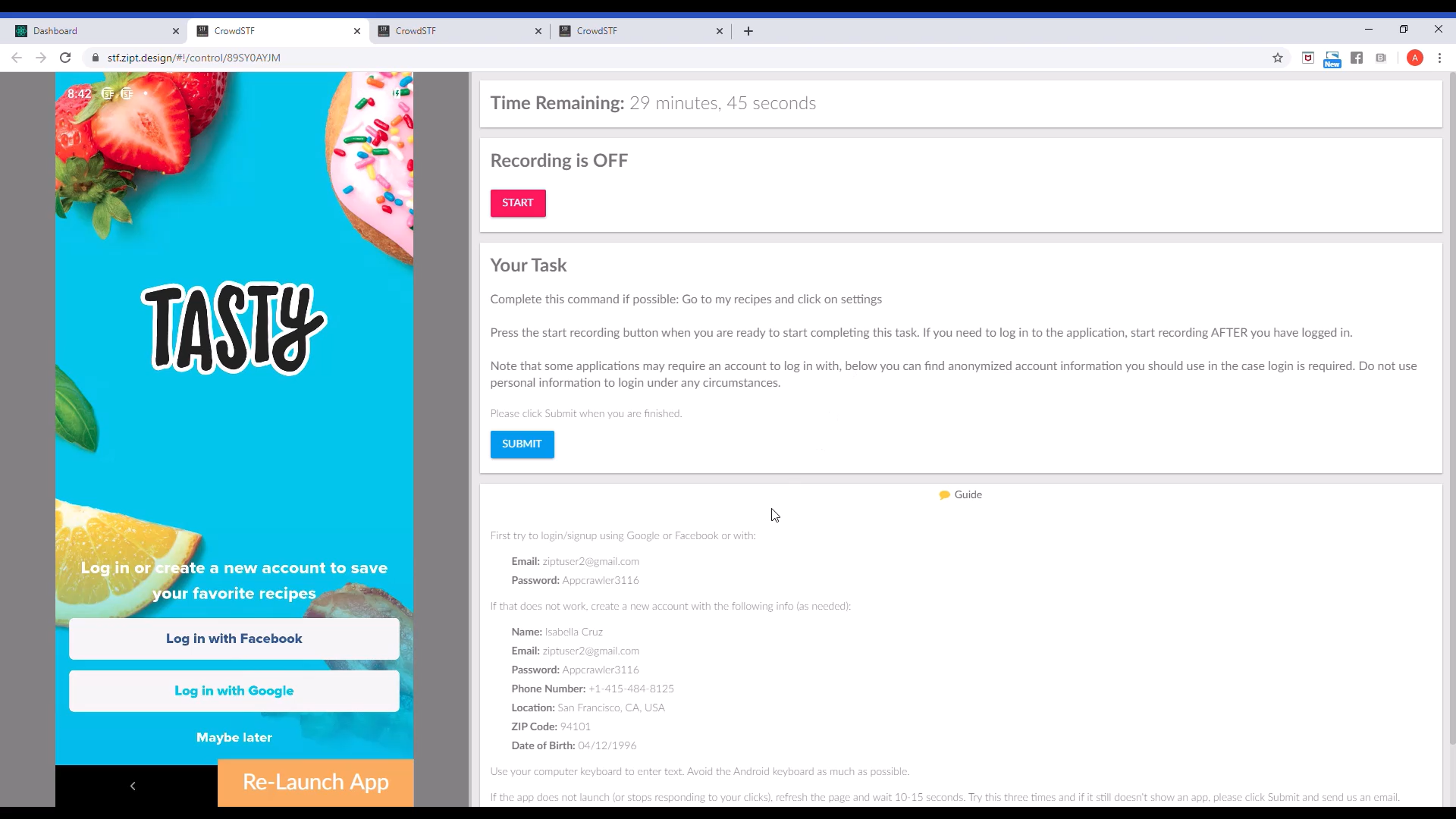}
    \caption{The website interface annotators use to interact with an Android app and record their task demonstration. We provide anonymized information if needed for logging in or for forms at any point so that no personal identifying information is collected}
    \label{fig:website}
\end{figure*}
We provide an example of what our data collection interface looks like for annotators while they explore an Android app and perform a task demonstration in Figure~\ref{fig:website}. Annotators are given the natural language task to attempt within the Android app in the `Your Task' section on the right side of the interface. Below, we provide anonymized email login and password credentials for them to use if needed. The left hand side of the collection interface displays the phone screen from a physical Android device which is remotely connected to our collection website, from which we record all actions taken on the phone and the app modalities as described in the main text.

\subsection{Application List}
We include lists of all Android apps we collect demonstrations for in Tables~\ref{tab:apps1}-\ref{tab:apps3}. In addition to listing the app package name, we provide the corresponding Google Play Store Category and how that particular app's tasks were paired (app-specific, paired, or category-clustered). The apps selected for MoTIF were across fifteen app categories: lifestyle, communication, dating, food and drink, maps and navigation, news and magazines, productivity, shopping, social, travel, weather, tools, music and audio, entertainment, and education. For privacy, we do not intend to collect any demonstrations of natural language commands within dating apps, and will not be releasing any of the raw data collected when annotators decided on a list of natural language tasks for dating apps in the first stage of collection. We simply include dating apps as one Android category to see what kinds of tasks people would consider being automated in this setting. We will share the resulting natural language tasks, but no captured screen or view hierarchy data. The dating apps included com.wildec.dating.meet4u, com.once.android, emotion.onekm, ru.fotostrana.sweetmeet, com.mason.wooplus, and com.hitwe.android.

\begin{table}
    \centering
        \caption{A list of applications used in MoTIF, their Google Play Store Category, and how their submitted natural language tasks were grouped with applications in the (app, task) pairing stage. N/A refers to apps which has technical difficulties during the demonstration stage and we are working to resolve}
    \begin{tabular}{|l|l|l|}
    \hline
         Google Play & \multirow{2}{*}{App Name} & (app, task) \\
         Store Category &  & Pairing Method \\
         \hline
         \multirow{9}{*}{Education} & com.ted.android & \textit{app-specific}\\
        \cline{2-3}
      &  gov.nasa & \textit{app-specific}\\ \cline{2-3}
       & example.matharithmetics & \textit{paired}\\ \cline{2-3}
       & org.khanacademy.android  & \textit{app-specific}\\ \cline{2-3}
      & com.duolingo  & \textit{app-specific} \\ \cline{2-3}
       & com.quizlet.quizletandroid & \textit{app-specific}\\ \cline{2-3}
       & com.remind101 & N/A \\ \cline{2-3}
       & org.coursera.android& N/A \\ \cline{2-3}
       & com.microblink.photomath & \textit{paired} \\ \hline
        \multirow{8}{*}{Entertainment} & com.megogo.application & \textit{app-specific} \\ \cline{2-3}
      &  com.app.emotes.dances.fortnite & \textit{app-specific}\\ \cline{2-3}
      &  com.scannerradio & \textit{app-specific}\\ \cline{2-3}
       & com.google.android.youtube & \textit{app-specific} \\ \cline{2-3}
       & com.zombodroid.MemeGenerator &\textit{app-specific} \\ \cline{2-3}
       & tv.pluto.android &\textit{app-specific} \\\cline{2-3}
       & com.tubitv & \textit{app-specific}\\\cline{2-3}
        & com.imdb.mobile & \textit{app-specific}\\\cline{2-3}
        & com.eventbrite.attendee & \textit{app-specific}\\ 
         \hline
         \multirow{6}{*}{Communication} & com.google.android.gm & \textit{app-specific} \\ \cline{2-3}
        & com.sec.android.app.sbrowser & \textit{paired}\\\cline{2-3}
        & com.facebook.orca & N/A \\ \cline{2-3}
        & com.whatsapp & N/A \\ \cline{2-3}
        & org.mozilla.firefox & \textit{paired} \\ \cline{2-3}
        & com.skype.raider & N/A \\ \hline
        \multirow{8}{*}{Food \& Drinks} & com.joelapenna.foursquared & \textit{app-specific} \\ \cline{2-3}
        & com.yum.pizzahut & \textit{app-specific} \\ \cline{2-3}
        & com.chickfila.cfaflagship&\textit{app-specific}  \\ \cline{2-3}
        & com.dominospizza & \textit{paired} \\ \cline{2-3}
        & in.swiggy.android &\textit{app-specific}  \\ \cline{2-3}
        & com.opentable & \textit{app-specific} \\ \cline{2-3}
        & com.starbucks.mobilecard & \textit{app-specific} \\ \cline{2-3}
       & vivino.web.app& \textit{app-specific} \\ \hline
         \multirow{7}{*}{Lifestyle} & com.hm.goe  & \textit{app-specific} \\ \cline{2-3}
        & com.adpog.diary  & \textit{app-specific} \\ \cline{2-3}
       &  com.aboutjsp.thedaybefore  & \textit{app-specific} \\ \cline{2-3}
       &  info.androidz.horoscope  & N/A \\ \cline{2-3}
        & ru.mail.horo.android & \textit{paired} \\ \cline{2-3}
       &  com.urbandroid.sleep & \textit{app-specific} \\ \cline{2-3}
        & com.hundred.qibla & \textit{app-specific}\\
         \hline
    \end{tabular}
    \label{tab:apps1}
\end{table}

\begin{table}
    \centering
        \caption{A list of applications used in MoTIF, their Google Play Store Category, and how their submitted natural language tasks were grouped with applications in the (app, task) pairing stage. N/A refers to apps which has technical difficulties during the demonstration stage and we are working to resolve}
    \begin{tabular}{|l|l|l|}
    \hline
         Google  & \multirow{3}{*}{App Name} & \multirow{2}{*}{(app, task)}  \\
         Play Store & & \multirow{2}{*}{Pairing Method} \\
         Category &  &  \\
         \hline
        & com.tranzmate & \textit{category-clustered} \\ \cline{2-3}
         & com.mapfactor.navigator& \textit{category-clustered}\\ \cline{2-3}
        Maps & com.thetrainline&\textit{category-clustered} \\ \cline{2-3}
       \& Navigation  & com.citymapper.app.release& \textit{app-specific}\\ \cline{2-3} &  com.prime.studio.apps.route.finder.map& \textit{category-clustered}\\ \cline{2-3}
         & com.waze& \textit{category-clustered}\\ \cline{2-3}
        & com.nyctrans.it &\textit{category-clustered} \\ \hline
          & com.radio.fmradio & \textit{app-specific}\\ \cline{2-3}
        & deezer.android.app&  \textit{app-specific}\\ \cline{2-3}
         & com.spotify.music& \textit{category-clustered}\\ \cline{2-3}
        Music & com.pandora.android& \textit{category-clustered}\\ \cline{2-3}
       \& Audio & com.springwalk.mediaconverter& \textit{category-clustered}\\ \cline{2-3} 
         & com.google.android.music& \textit{category-clustered}\\ \cline{2-3}
         & com.clearchannel.iheartradio.controller& \textit{category-clustered}\\ \cline{2-3}
        & com.melodis.midimiMusicIdentifier.freemium & \textit{category-clustered}\\
         \hline
         &  fm.castbox.audiobook.radio.podcast& \textit{category-clustered} \\ \cline{2-3}
         & com.ss.android.article.master& N/A \\ \cline{2-3}
         & com.opera.app.news& \textit{category-clustered}\\ \cline{2-3}
        News  & bbc.mobile.news.ww& \textit{category-clustered}\\\cline{2-3}
       \& Magazines  & com.quora.android&  N/A \\ \cline{2-3}
         & com.google.android.apps.magazines& \textit{category-clustered} \\ \cline{2-3}
         & com.reddit.frontpage& \textit{app-specific}\\ \cline{2-3}
         & com.sony.nfx.app.sfrc & \textit{category-clustered} \\
         \hline 
        \multirow{7}{*}{Shopping} & com.amazon.mShop.android.shopping& \textit{app-specific}\\ \cline{2-3}
         & com.abtnprojects.ambatana&  \textit{category-clustered}\\ \cline{2-3}
         & com.contextlogic.wish& \textit{category-clustered}\\ \cline{2-3}
         & com.joom& \textit{category-clustered} \\ \cline{2-3}
          & com.ebay.mobile& \textit{category-clustered}\\ \cline{2-3}
         & com.walmart.android& \textit{category-clustered} \\ \cline{2-3}
         & club.fromfactory& \textit{app-specific}\\ \cline{2-3}
         & com.zzkko& \textit{app-specific}\\ \cline{2-3}
         & com.groupon&  \textit{category-clustered}\\
         \hline
         \multirow{7}{*}{Productivity} & cn.wps.moffice\_eng& \textit{category-clustered}\\ \cline{2-3}
         & com.google.android.apps.docs.editors.sheets& \textit{category-clustered}\\ \cline{2-3}
         & com.google.android.apps.docs& N/A \\ \cline{2-3}
         & com.microsoft.office.outlook& \textit{category-clustered}\\ \cline{2-3}
         & com.google.android.calendar &\textit{category-clustered} \\ \cline{2-3}
          & com.google.android.apps.docs.editors.slides& \textit{category-clustered}\\ \cline{2-3}
          & com.dropbox.android & N/A \\ \hline
    \end{tabular}
    \label{tab:apps2}
\end{table}

\begin{table}
    \centering
        \caption{A list of applications used in MoTIF, their Google Play Store Category, and how their submitted natural language tasks were grouped with applications in the (app, task) pairing stage. N/A refers to apps which has technical difficulties during the demonstration stage and we are working to resolve}
    \begin{tabular}{|l|l|l|}
    \hline
         Google  & \multirow{3}{*}{App Name} & \multirow{2}{*}{(app, task)}  \\
         Play Store & & \multirow{2}{*}{Pairing Method} \\
         Category &  &  \\
         \hline
          \multirow{7}{*}{Tools}&  com.lenovo.anyshare.gps& \textit{app-specific}\\ \cline{2-3}
         & com.antivirus& \textit{paired} \\ \cline{2-3}
       &  com.google.android.calculator& \textit{paired}\\ \cline{2-3}
         & com.miui.calculator& \textit{paired}\\ \cline{2-3}
        & com.google.android.apps.translate& \textit{app-specific}\\ \cline{2-3}
       & com.avast.android.mobilesecurity & \textit{paired}\\
       \hline
       \multirow{9}{*}{Travel}&  com.kayak.android& \textit{paired} \\ \cline{2-3}
         & com.tripadvisor.tripadvisor& \textit{paired}\\ \cline{2-3}
         & com.trivago& \textit{paired}\\ \cline{2-3}
         & com.google.android.apps.maps& \textit{paired}\\ \cline{2-3}
         & com.yelp.android& \textit{app-specific}\\ \cline{2-3}
         & com.booking& N/A \\ \cline{2-3}
         & com.google.earth& \textit{paired} \\ \cline{2-3}
         & com.mapswithme.maps.pro& \textit{app-specific}\\ \cline{2-3}
         & com.google.android.street& \textit{paired}\\ \cline{2-3}
        & com.yellowpages.android.ypmobile & \textit{app-specific}\\
       \hline
      \multirow{9}{*}{Weather} &  com.gau.go.launcherex.gowidget.weatherwidget& N/A \\ \cline{2-3}
          & com.devexpert.weather& \textit{category-clustered} \\ \cline{2-3}
          & com.chanel.weather.forecast.accu & \textit{category-clustered}\\ \cline{2-3}
         & com.weather.Weather& \textit{category-clustered}\\ \cline{2-3}
          & com.droid27.transparentclockweather& \textit{app-specific}\\ \cline{2-3}
         & aplicacion.tiempo & \textit{category-clustered}\\ \cline{2-3}
         & com.accuweather.android & \textit{category-clustered}\\ \cline{2-3}
         & com.windyty.android & \textit{category-clustered}\\ \cline{2-3}
         & com.handmark.expressweather & \textit{category-clustered}\\
       \hline
       \multirow{7}{*}{Social}&  com.zhiliaoapp.musically & \textit{category-clustered} \\ \cline{2-3}
          & com.pinterest & \textit{category-clustered} \\ \cline{2-3}
           & com.instagram.android & \textit{category-clustered}\\ \cline{2-3}
          & com.facebook.katana & \textit{category-clustered} \\ \cline{2-3}
           & com.sgiggle.production & \textit{app-specific}\\ \cline{2-3}
           & com.snapchat.android & \textit{app-specific}\\ \cline{2-3}
           & com.ss.android.ugc.boom & \textit{category-clustered} \\ \cline{2-3}
           & com.lazygeniouz.saveit & \textit{category-clustered}\\ \hline
    \end{tabular}
    \label{tab:apps3}
\end{table}

\subsection{Dataset Examples}
\begin{figure}[t]
    \centering
    \includegraphics[scale=0.2]{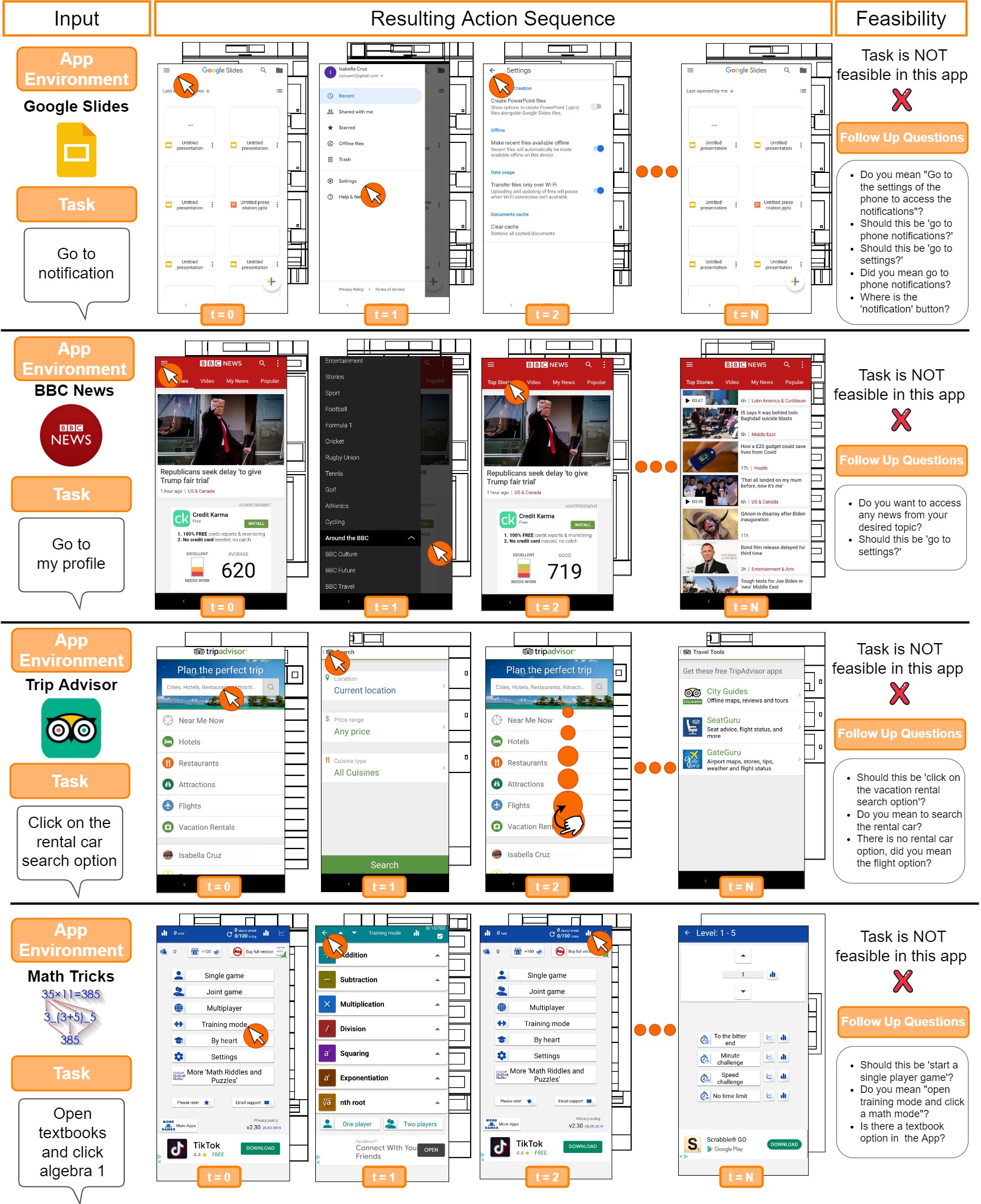}
            \caption{Example tasks from MoTIF deemed infeasible by annotators. We show the input (app, task) pair for task demonstration, the resulting task demo (which captures the rendered screen, app view hierarchy, and action localization), and the feasibility annotations and follow up questions posed by annotators}
                \label{fig:infeas_motif_ex}
\end{figure}

\begin{figure}[t]
    \centering

    \includegraphics[scale=0.2]{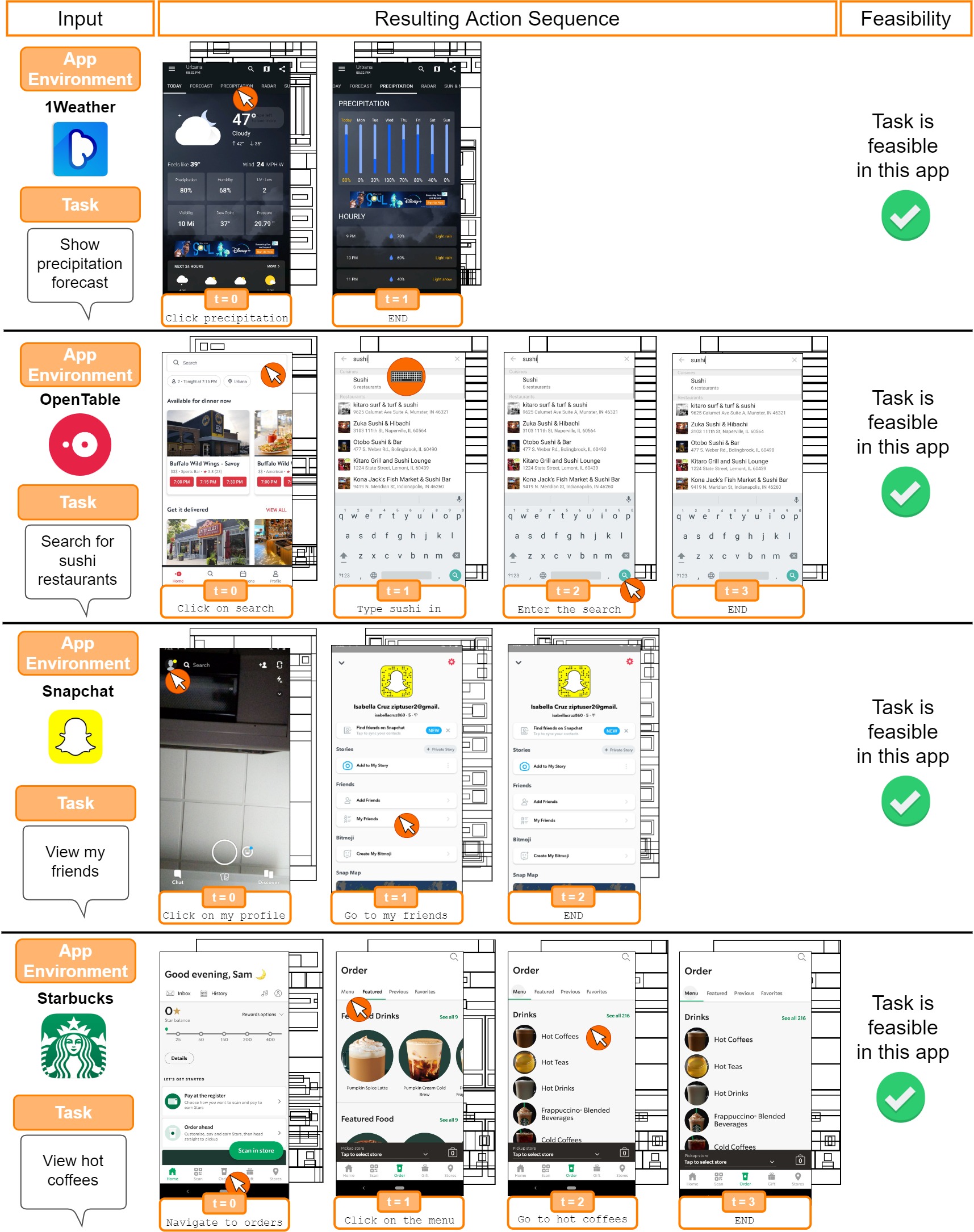}
            \caption{Example tasks from MoTIF deemed feasible by annotators. We show the input (app, task) pair for task demonstration, the resulting task demo (which captures the rendered screen, app view hierarchy, and action localization), and the feasibility annotations and follow up questions posed by annotators}
    \label{fig:feas_motif_ex}
\end{figure}

We include more example (app, task) pairs and their resulting action sequences from MoTIF. Figure~\ref{fig:infeas_motif_ex} and~\ref{fig:feas_motif_ex} show samples for infeasible and feasible commands, respectively.

\clearpage

\section{MoTIF Statistics}
We include statistics over the high-level goals collected for MoTIF in Section~\ref{sec:nl_stats} and word cloud visualizations over all commands and per category in Section~\ref{sec:wordcloud}. We discuss annotator agreement when determining command feasibility in Section~\ref{sec:agree}. Lastly, the cluster visualizations used to define (app, task) pairs in MoTIF are illustrated in Section~\ref{sec:tsne}.

\label{sec:stats}
\begin{figure}[t]
\centering
\begin{subfigure}{0.48\textwidth}
  \centering
  \includegraphics[width=1.0\linewidth]{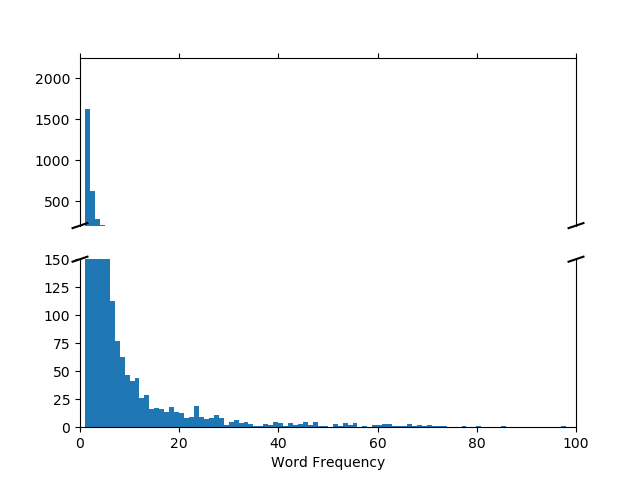}
  \caption{The word frequency distribution of MoTIF's vocabulary}
  \label{fig:wordfreq}
\end{subfigure}
\begin{subfigure}{0.48\textwidth}
  \centering
  \includegraphics[width=1.0\linewidth]{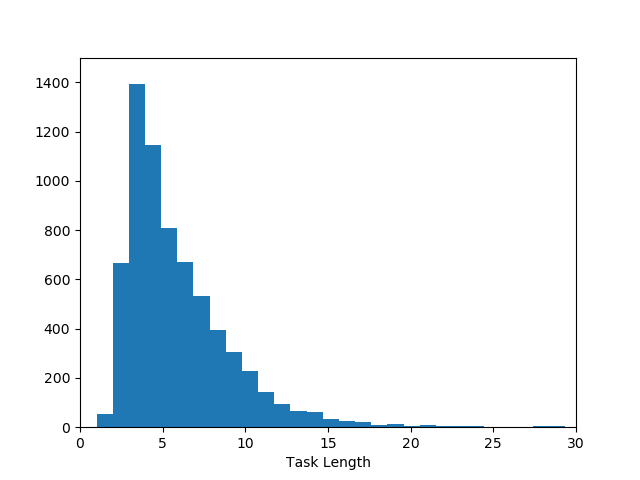}
  \caption{The length (number of words per task) distribution of MoTIF's tasks}
  \label{fig:tasklen}
\end{subfigure}
\caption{Additional statistics on MoTIF's language tasks}
\label{fig:langanalysis}
\end{figure}

\subsection{Natural Language Command Statistics}
\label{sec:nl_stats}
We provide additional statistics on the natural language high-level goals in MoTIF in Figure~\ref{fig:langanalysis}. In Figure~\ref{fig:wordfreq} we plot a histogram over the word frequency of the command vocabulary and Figure~\ref{fig:tasklen} shows a histogram over the task length (\ie, how many words a task consists of) across all collected natural language tasks. Both reflect a long tail distribution, which is common for word frequency, and follows Zipf's Law. For task length, the distribution is skewed towards shorter length tasks (nearly all collected tasks have fewer than ten words), which aligns with MoTIF's natural language commands mostly capturing high-level goals.
\subsection{Word Cloud Visualizations}
\label{sec:wordcloud}
\begin{figure}[t]
    \centering
    \includegraphics[scale=0.05]{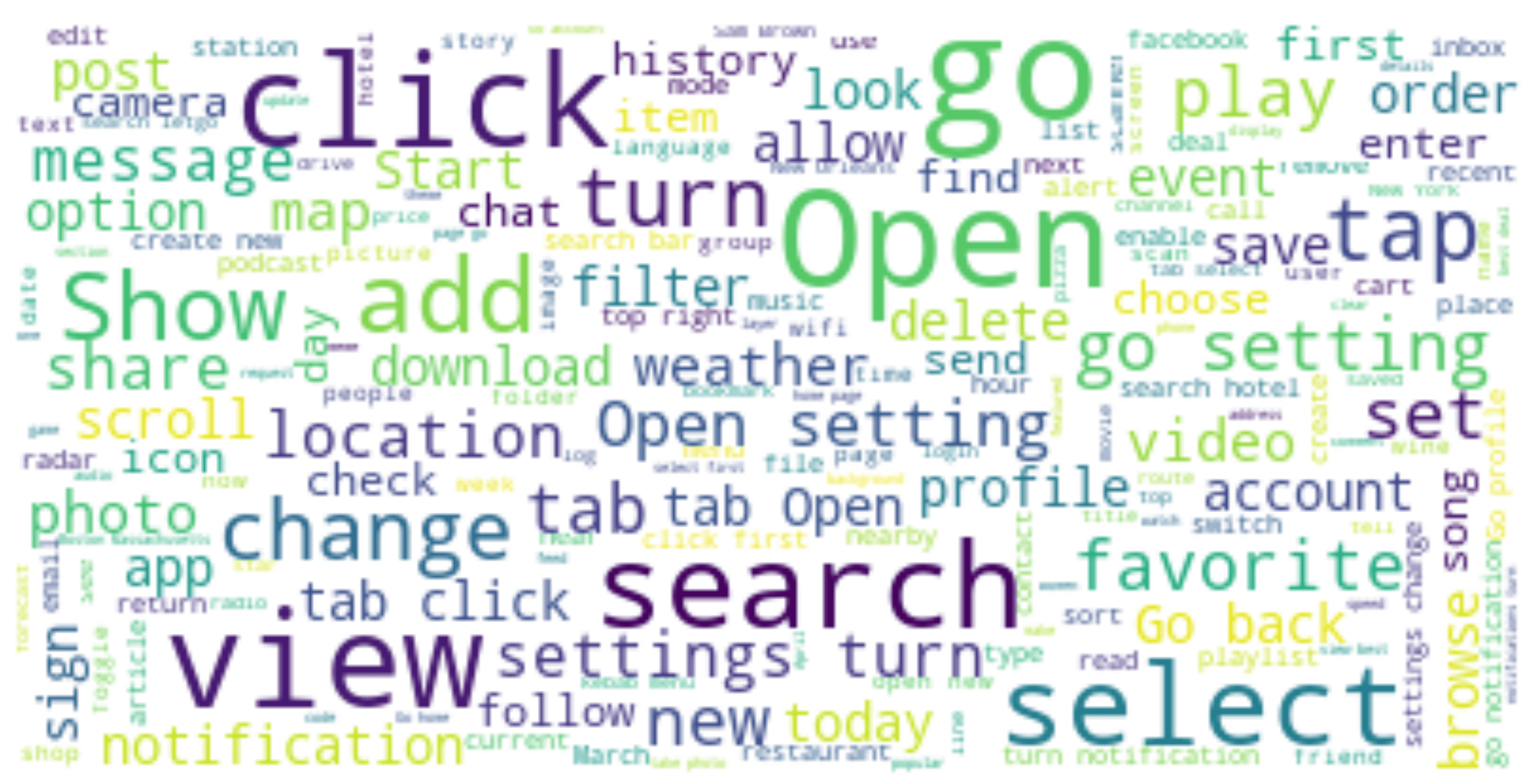}
    \caption{Word cloud visualization over all MoTIF high-level language commands. The larger the word is illustrated, the more often it occurs}
    \label{fig:word1}
\end{figure}

\begin{figure}
    \centering
    \includegraphics[scale=0.209]{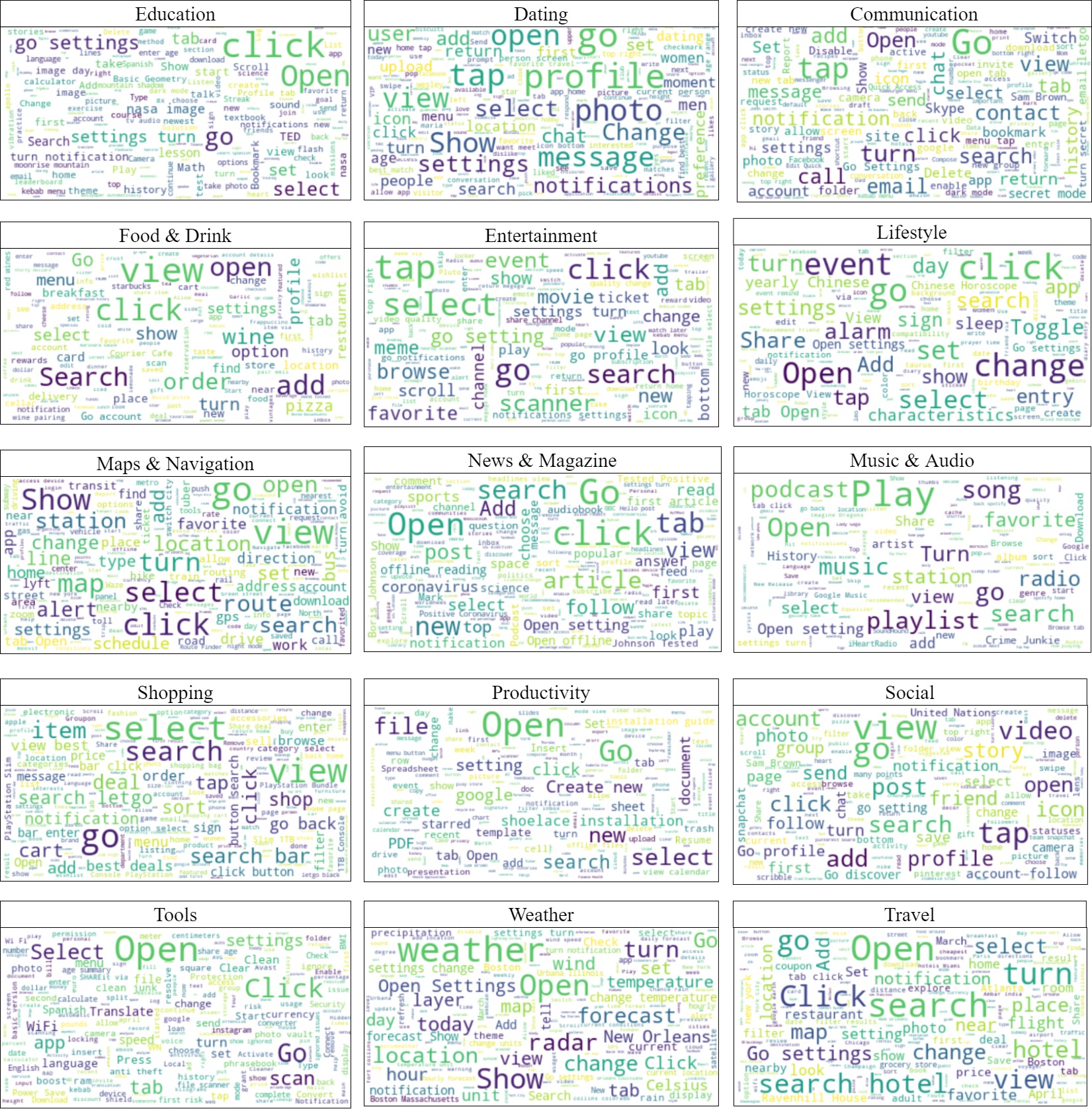}
    \caption{Word cloud visualization of MoTIF high-level language tasks per Android app category. There are fifteen total categories: Education, Dating, Communication, Food \& Drink, Entertainment, Lifestyle, Maps \& Navigation, News \& Magazine, Music \& Audio, Shopping, Productivity, Social, Tools, Weather, and Travel. The larger the word is illustrated, the more often it occurs}
    \label{fig:category_wordcloud}
\end{figure}

We include a word cloud illustration over all high-level commands in MoTIF in Figure~\ref{fig:word1}. The larger the word in the word cloud, the more often it occurs in MoTIF's collected tasks. As we compute the word cloud over all tasks (which span fifteen different Google Play Store app categories) we can see the largest words are those that are action or instruction oriented words, like `click,' `search,' or `show.' In Figure~\ref{fig:category_wordcloud}, we show word clouds for tasks per app category. 

While there are some common words with high frequency across all app categories (like the action oriented words largest in Figure~\ref{fig:word1}), there are other words illustrated that reflect each app category and functionality specific to that topic. For example, in the Education word cloud in the top left of Figure~\ref{fig:category_wordcloud}, we see words `lesson,' `math,' and `history.' In contrast, the Shopping category in Figure~\ref{fig:category_wordcloud} shows words like `deal,' `search,' and `cart' with high frequency.

The word cloud visualizations also show the density of words for each Android app category's collected tasks. The Food \& Drink, Productivity, and Music \& Audio app categories have the smallest vocabularies, with less densely populated word clouds. This reflects there being lower diversity in the kinds of requests asked by people for these app categories. On the other hand, Maps \& Navigation, Weather, and Travel are examples of Android app categories with larger task vocabularies. This can reflect greater diversity in app requests collected, which may be due to the diversity of functionality in these app categories, or the fact that these apps can have highly specific, \ie, very fine-grained, requests (like searching for one location's weather out of the nearly unlimited locations one could request). 

\subsection{Annotator Feasibility Agreement}
\label{sec:agree}
\begin{figure}[t]
    \centering
    \includegraphics[scale=0.55]{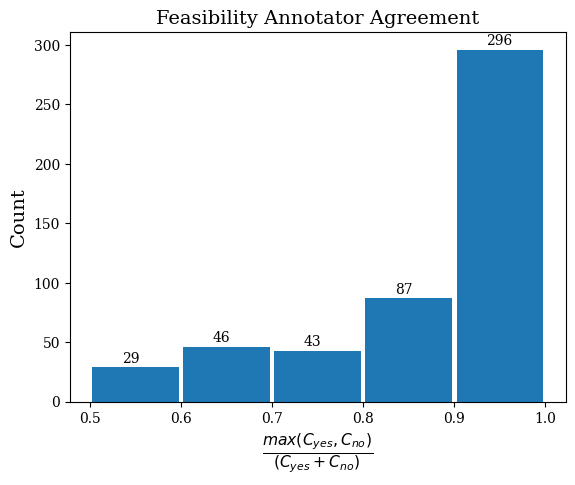}
    \caption{The annotator feasibility labeling agreement for (app, task) pairs with demonstrations in MoTIF}
    \label{fig:agree}
\end{figure}

We define annotator feasibility labeling agreement as the fraction of the number of votes for the majority voted label ($max(C_{yes}, C_{no})$) over all votes ($C_{yes} + C_{no}$) for an (app, task) pair in MoTIF, where $C_{yes}$ is the count of votes for feasible and $C_{no}$ is the count of votes for infeasible. In Figure~\ref{fig:agree}, we bin different degrees of annotator agreement and plot each bin's counts over all (app, task) pairs with demonstrations in MoTIF. The minimum agreement is 50\% and maximum agreement is 100\%. The majority of our (app, task) pairs have annotation agreement between 90-100\%, with 296 (app, task) pairs falling in this maximal bin.

\subsection{App Category Clustering Visualizations}
\label{sec:tsne}
\begin{figure}
    \centering
    \includegraphics[scale=0.085]{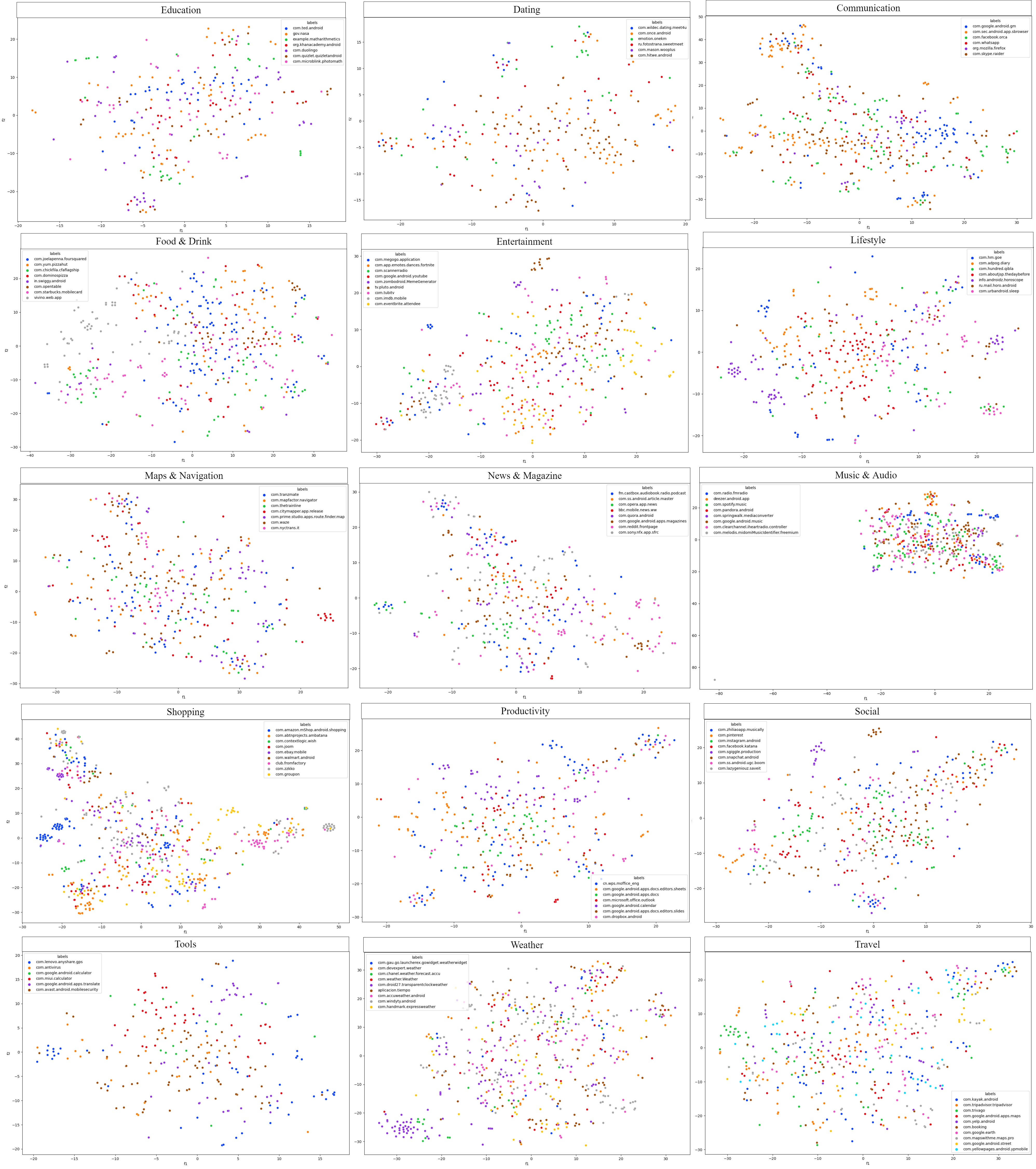}
    \caption{T-SNE visualization of K-Means clusters for each Android Google Play Store Category. The visualizations are colored with the originating app label (and not the K-Means cluster label). These visualizations are used to inspect which apps should retain
    their app-specific tasks during the action sequence demonstration stage}
    \label{fig:all_app_tsne}
\end{figure}

\begin{figure}
    \centering
    \includegraphics[scale=0.1325]{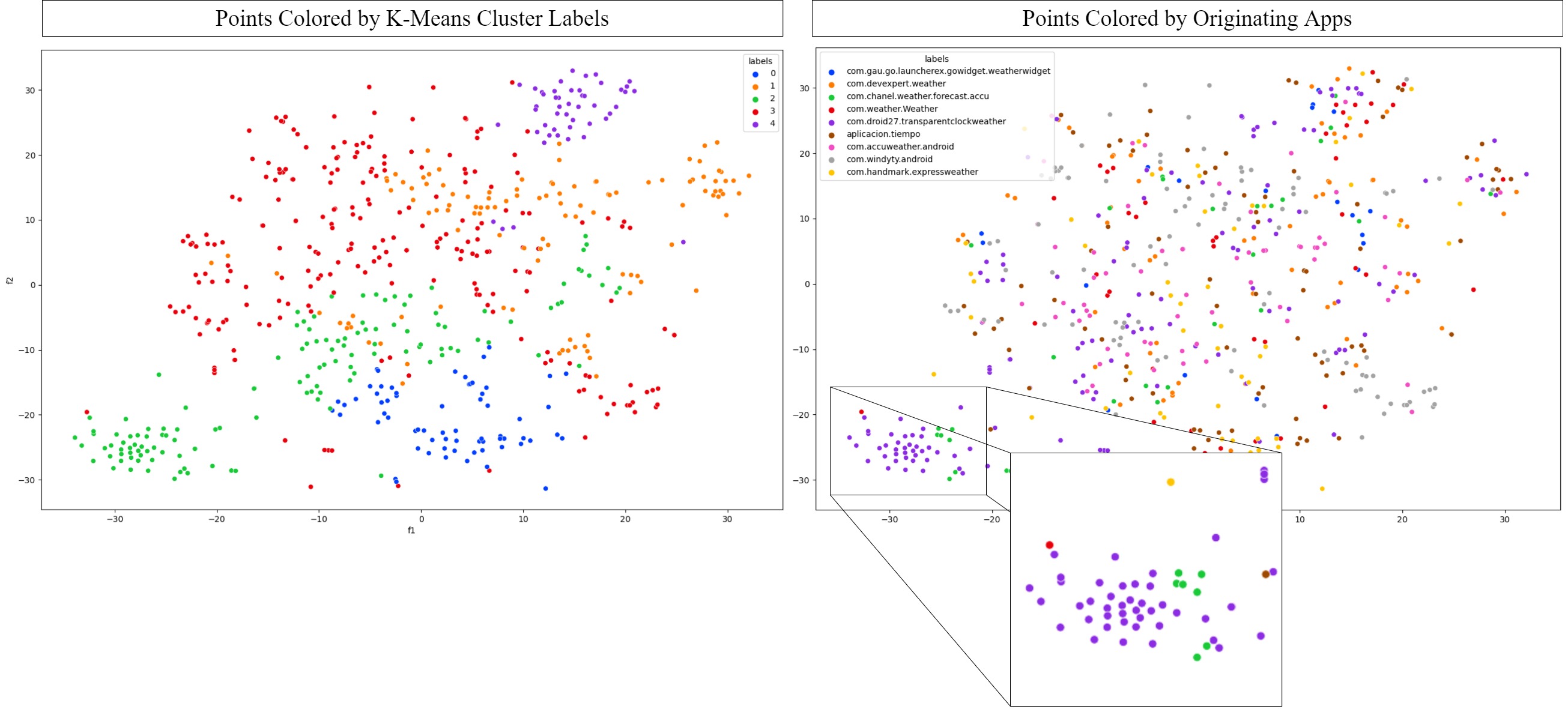}
    \caption{T-SNE visualization of K-Means clusters on MoTIF commands from the Weather Google Play Store app category. Points represent MoTIF commands (represented by their mean FastText embedding). The left plot colors points by the clusters output by K-Means, while the right plot colors points by their originating app. In the lower left corner of both plots is a cluster (the green cluster on the left hand side), which when colored by the app the command was originally written for (on the right hand side), we see primarily comes from a single app, com.droid27.transparentclockweather. As a result, this app's commands will not be category clustered, and will stay paired with com.droid27.transparentclockweather}
    \label{fig:tsne1}
\end{figure}

We provide the K-Means T-SNE cluster visualizations used in the (app, task) pairing process for each category of apps in Figure~\ref{fig:all_app_tsne}. These clusters decide whether an app's tasks are kept app-specific, paired to one or two other apps, or are category clustered. We zoom into the cluster visualization for the Weather Android app category in Figure~\ref{fig:tsne1}. On the left, we see the cluster output for K-Means on the average task embedding (using FastText representations) for the commands written for weather apps. On the right we show the exact same clustering, but now color the points (\ie, the written tasks) by which app they were originally written for. In the lower left corner of the cluster visualization is an isolated cluster for the com.droid27.transparentclockweather app. As its tasks form an isolated cluster, they are kept app-specific, while all other apps have (app, task) pairs obtained from the category clustering. 

To actually select the category clustered tasks, we select natural language commands near each cluster's centroid. These serve as cluster representatives for our task demonstration data collection. So, for every Google Play Store app category, we perform K-Means with K=5, as we start by collecting demonstrations for five commands per app. Then, for apps that are chosen to be category clustered, we select the cluster representatives and collect demonstrations of these representatives for each weather app. For additional clarity, see Tables~\ref{tab:apps1}-\ref{tab:apps3} for the (app, task) pairing method per app. Eventually, the goal is to collect all possible combinations of (app, task) pairs within a category.

\begin{table}[t]
    \centering
        \caption{Task feasibility F1 score using our MLP. We ablate input features and how action demonstration sequences are aggregated. The random baseline predicts a feasibility label given the train set distribution}
    \begin{tabular}{|l|c|c|c|}
    \hline
      \multirow{2}{*}{\textbf{C}$_{feas}$ Input Features} & \multicolumn{3}{c|}{Demo Aggregation}\\
      \cline{2-4}
      & Avg & Cat & LSTM \\ 
      \hline
      \textbf{Random} & \multicolumn{3}{c|}{20.1}\\
      \hline
     \textbf{(a) View Hierarchy} & & & \\ 
     FastText & & &  \\
    \hspace{4mm} ET & 22.8 & 44.3 & 37.0 \\
    \hspace{4mm} ET + ID & 16.7 & 43.6 & 34.1 \\ 
     
    \hspace{4mm} ET + ID + CLS & 19.7 & 39.6 & 36.2 \\
    CLIP & &  & \\
    \hspace{4mm} ET & 27.0 & 48.4 & 35.9 \\
     \hspace{4mm} ET + ID & 28.0 & 50.9 & 36.2 
     \\
    \hspace{4mm} ET + ID + CLS & 29.6 & 49.2 & 35.2  \\
      Screen2Vec & 25.9 & 33.7 & 36.0 \\ 
      \hline
      \textbf{(b) App Screen Image} & & & \\ 
      ResNet & 31.3 & 41.9 & 35.9 \\ 
     Icons & 0.4 & 40.0 & 15.2 \\
     CLIP & 44.7 & 58.2 & \underline{42.8} \\
      \hline
      \textbf{(c) Best Combination} & & & \\
      CLIP (Screen + ET + ID) & \underline{44.8} & \underline{61.1} & 40.9 \\ 
      \hline
    \end{tabular}

    \label{tab:full_feas}
\end{table} 

\section{Task Feasibility Experiments}
\label{sec:feas}
In Table~\ref{tab:full_feas}(a), we have additional rows for which view hierarchy element attributes are included as input features to our feasibility classifier. The view hierarchy of an Android app contains several element attributes, including text (ET), resource-identifier (ID), and class (CLS) attributes. We ablate using one or multiple of these attributes and find that on average across demonstration aggregation type, the (ET + ID) input combination results in the best performance. Consequently, we keep it for our best results in the main text.

\section{Task Automation Experiments}
We further detail how task automation experiments are performed in a vision-language navigation paradigm in Section~\ref{sec:app_graph}, where we describe the test-time environment. Then, we report performance when training VLN methods only on our data in Section~\ref{sec:auto_data}. In Section~\ref{sec:auto_language}, we evaluate our models from the main paper on different language inputs (high-level goal, low-level instruction, or both) at test-time and describe performance trends. Lastly, in Section~\ref{sec:task_gen} we include some additional results on generalization of tasks across apps for a subset of our baselines. 

\subsection{Test-time Evaluation of Seq2Seq and MOCA}
\label{sec:app_graph}
We build an offline version of each Android app environment to approximate a complete state-action space graph at test time. We merge demonstrations we've collected across all samples.
The nodes in this state-action space graph are unique `views' of an application, \ie, a particular screen within an action demonstration sequence. Nodes are connected by edges which represent the transition between any pair of screens. This transition is defined by the action class (clicking, typing, or swiping) and the location of the action taken at the current screen state (point or bounding box coordinates in the rendered app screen image).
\begin{table}[t]
    \centering
          \renewcommand\arraystretch{0.95}

        \caption{Mobile app task complete and partial sequence accuracy on MoTIF when trained on MoTIF alone, or MoTIF and RicoSCA data for the Seq2Seq model. The training and testing language input are kept the same; input contains the high-level goal and low level step by step instructions }
    \begin{tabular}{|l|c|c|c|c|c|c|c|}
    \hline
    & & \multicolumn{6}{c|}{MoTIF Test Split}\\
    \cline{3-8}
      Model & Train & \multicolumn{3}{c|}{App Seen} & \multicolumn{3}{c|}{App Unseen}\\
    \cline{3-8}
 \textbf{Seq2Seq} & Data & \multirow{2}{*}{Action} & \multirow{2}{*}{Ground} & Action + &  \multirow{2}{*}{Action} & \multirow{2}{*}{Ground} & Action +\\
   & & & & Ground& & & Ground\\
    \hline
    Complete & \multirow{2}{*}{MoTIF} & 45.0 & 17.1 & 15.9 & 33.8 & 13.6 & 11.7 \\
    Partial & & 79.4 & 37.7 & 35.5 & 66.8 & 27.8 & 25.0 \\
    \hline
    Complete & MoTIF + & 68.5 & 22.5 & 22.5 & 54.3 & 18.0 & 17.7 \\
    Partial & RicoSCA & 89.5 & 40.4 & 40.1 & 81.7 & 31.3 & 30.6 \\
    \hline
    \end{tabular}
    \label{tab:automate_full1}
\end{table}

\begin{table}[t]
    \centering
          \renewcommand\arraystretch{0.95}

        \caption{Mobile app task complete and partial sequence accuracy on MoTIF when trained on MoTIF alone, or MoTIF and RicoSCA data for the MOCA model. The training and testing language input are kept the same; input contains the high-level goal and low level step by step instructions }
    \begin{tabular}{|l|c|c|c|c|c|c|c|}
    \hline
    & & \multicolumn{6}{c|}{MoTIF Test Split}\\
    \cline{3-8}
      Model & Train & \multicolumn{3}{c|}{App Seen} & \multicolumn{3}{c|}{App Unseen}\\
    \cline{3-8}
   \textbf{MOCA} & Data & \multirow{2}{*}{Action} & \multirow{2}{*}{Ground} & Action + &  \multirow{2}{*}{Action} & \multirow{2}{*}{Ground} & Action +\\
   & & & & Ground& & & Ground\\
    \hline
    Complete & \multirow{2}{*}{MoTIF} & 37.8 & 16.2 & 12.3 & 24.6 & 17.0 & 13.2 \\
    Partial & & 66.0 & 34.9 & 29.9 & 60.4 & 32.0 & 27.7 \\
    \hline
    Complete & MoTIF +  & 51.1 & 21.3 & 20.7 & 44.8 & 17.0 & 15.1 \\
    Partial & RicoSCA & 78.5 & 40.0 & 38.6 & 72.2 & 32.7 & 30.0 \\
    \hline
    \end{tabular}
    \label{tab:automate_full2}
\end{table}
\subsection{Training Data Ablations}
\label{sec:auto_data}

We also ran experiments with Seq2Seq and MOCA when trained only on MoTIF data instead of both MoTIF and RicoSCA. We include these comparisons for Seq2Seq and MOCA in Tables~\ref{tab:automate_full1} and~\ref{tab:automate_full2}, respectively. Jointly training on both datasets consistently performs better across all metrics. Additionally, performance trends generally remain the same when comparing the app seen versus app unseen test split: regardless of training data, accuracy is higher on the app seen test split.
We report the joint training performance for these methods in the main text for a closer apples-to-apples comparison with Seq2Act.

\begin{table}[t]
    \centering
          \renewcommand\arraystretch{0.95}

        \caption{Mobile app task complete and partial sequence accuracy on MoTIF with various language inputs at test time for the Seq2Seq model. The training input contains the high level goal and low level step by step instructions }
    \begin{tabular}{|l|c|c|c|c|c|c|c|}
    \hline
    & & \multicolumn{6}{c|}{MoTIF Test Split}\\
    \cline{3-8}
      Model & Test & \multicolumn{3}{c|}{App Seen} & \multicolumn{3}{c|}{App Unseen}\\
    \cline{3-8}
  \textbf{Seq2Seq} & Input & \multirow{2}{*}{Action} & \multirow{2}{*}{Ground} & Action + &  \multirow{2}{*}{Action} & \multirow{2}{*}{Ground} & Action +\\
   & & & & Ground& & & Ground\\
    \hline
    Complete & High + & 68.5 & 22.5 & 22.5 & 54.3 & 18.0 & 17.7 \\
    Partial & Low & 89.5 & 40.4 & 40.1 & 81.7 & 31.3 & 30.6 \\
    \hline
    Complete &\multirow{2}{*}{Low} & 47.1 & 18.6 & 18.0 & 27.1 & 13.9 & 13.9 \\
    Partial & & 73.7 & 36.6 & 33.9 & 43.6 & 22.6 & 21.2 \\
    \hline
    Complete & \multirow{2}{*}{High} & 30.9 & 15.3 & 14.7 & 18.9 & 11.7 & 8.8 \\
    Partial & & 68.1 & 31.6 & 29.5 & 59.1 & 24.0 & 19.8 \\
    \hline
    \end{tabular}
    \label{tab:automate_eval1}
\end{table}
\subsection{Test-time Language Input Ablations}
\label{sec:auto_language}
We include ablations for the trained models in the main text for all possible language inputs at test time. Seq2Seq and MOCA were trained on both high-level goal and low-level instructions, as their original models supported both inputs and obtained best performance with them in prior work. Seq2Act does not currently support high-level goal language input, so we cannot jointly evaluate both in a meaningful way. We benchmark models as close to their original architecture as possible, and leave adaptations to future work.

In the main text, all task automation results were reported on the same language input as was used during training to avoid confounding factors when analyzing generalization to new app environments. Thus, Seq2Seq and MOCA took both high-level and low-level command as input while Seq2Act took only low-level instruction. We now evaluate all possible input language ablations at test time. Evaluating the high-level goal input alone replicates what these models would be provided in practical application, as users would request high-level goals (and not provide step by step instruction). Our high-level input results are useful to evaluate generalization to downstream settings, but we also include results for low-level input alone or both high-level and low-level language instruction (where applicable, as Seq2Act cannot support both) in Tables~\ref{tab:automate_eval1},~\ref{tab:automate_eval2}, and~\ref{tab:automate_eval3}.

The Seq2Seq partial and complete sequence accuracy for action prediction show that having both high-level goal and low-level instruction inputs result in the best performance, followed by low-level instruction, and then high-level goal. On the other hand, MOCA performs quite similarly when both high-level goal and low-level instruction are input versus low-level instruction alone on action prediction. Additionally, there is less grounding performance degradation over the ablations, which may be a result of MOCA's more constrained test-time environment (which uses app type prediction to narrow the grounding prediction space).

Seq2Act performs best across all metrics when provided the low-level instruction at test time. This is expected, given that Seq2Act was trained on step by step instructions. For both test splits, the action and grounding accuracy is significantly higher with low-level input. As the VLN methods showed having both high-level and low-level inputs can improve performance, adapting Seq2Act to take both as input would be important in future work.

\begin{table}[t]
    \centering
          \renewcommand\arraystretch{0.95}

        \caption{Mobile app task complete and partial sequence accuracy on MoTIF with various language inputs at test time for the MOCA model. The training input contains the high level goal and low level step by step instructions }
    \begin{tabular}{|l|c|c|c|c|c|c|c|}
    \hline
    & & \multicolumn{6}{c|}{MoTIF Test Split}\\
    \cline{3-8}
      Model & Test & \multicolumn{3}{c|}{App Seen} & \multicolumn{3}{c|}{App Unseen}\\
    \cline{3-8}
  \textbf{MOCA} & Input & \multirow{2}{*}{Action} & \multirow{2}{*}{Ground} & Action + &  \multirow{2}{*}{Action} & \multirow{2}{*}{Ground} & Action +\\
   & & & & Ground& & & Ground\\
    \hline
    Complete & High + & 51.1 & 21.3 & 20.7 & 44.8 & 17.0 & 15.1 \\ 
    Partial & Low & 78.5 & 40.0 & 38.6 & 72.2 & 32.7 & 30.0 \\
    \hline
    Complete &\multirow{2}{*}{Low} & 48.6 & 19.5 & 19.2 & 45.4 & 17.0 & 15.8\\
    Partial & & 77.3 & 36.5 & 36.5 & 74.1 & 32.4 & 30.8 \\ 
    \hline
    Complete & \multirow{2}{*}{High}&  13.5 & 19.5 & 8.4 & 11.4 & 18.6 & 6.9\\ 
    Partial & & 43.6 & 38.8 & 26.1 & 41.1 & 33.5 &21.2 \\
    \hline
    \end{tabular}
    \label{tab:automate_eval2}
\end{table}

\begin{table}[t]
    \centering
          \renewcommand\arraystretch{0.95}

        \caption{Mobile app task complete and partial sequence accuracy on MoTIF with various language inputs at test time for the Seq2Act model. The training input contains the low level step by step instructions }
    \begin{tabular}{|l|c|c|c|c|c|c|c|}
    \hline
    & & \multicolumn{6}{c|}{MoTIF Test Split}\\
    \cline{3-8}
      Model & Test & \multicolumn{3}{c|}{App Seen} & \multicolumn{3}{c|}{App Unseen}\\
    \cline{3-8}
  \textbf{Seq2Act} & Input & \multirow{2}{*}{Action} & \multirow{2}{*}{Ground} & Action + &  \multirow{2}{*}{Action} & \multirow{2}{*}{Ground} & Action +\\
   & & & & Ground& & & Ground\\
    \hline
    Complete &\multirow{2}{*}{Low} & 97.3 & 32.4 & 32.4 & 96.8 & 28.3 & 28.3 \\
    Partial & & 99.2 & 66.4 & 66.3 & 99.6 & 67.7 & 67.6 \\
    \hline
    Complete & \multirow{2}{*}{High}&  10.6 & 7.6 & 7.6 & 8.5 &1.9&1.9\\
    Partial & & 28.1 & 13.0  & 10.8 &31.3 & 7.0 & 5.4\\
    \hline
    \end{tabular}
    \label{tab:automate_eval3}
\end{table}

\subsection{Generalization of Natural Language Commands across Apps}
\label{sec:task_gen}
We lastly evaluate generalization of our task automation methods to natural language tasks. Specifically, we present results on two additional test splits: an app seen and task unseen app split (where the task was seen in other apps, but not the current) and an app unseen and task seen split. The former shows the easier setting of having seen the app environment with other tasks during training and the task with other apps during training, whereas the app unseen test split means the task was seen during training with other apps but the model has never seen any task in this particular app.

Intuitively, performance is consistently higher on the easier setting of app seen and task unseen (current app), as the model has had the chance to learn about both the app environment and task instruction, albeit independently. Comparing these task generalization results to the app generalization results in the main text (can also be found in Tables~\ref{tab:automate_eval1}-\ref{tab:automate_eval3}), the models can consistently generalize tasks across applications better than they can generalize to new environments. 

\begin{table}[t]
    \centering
          \renewcommand\arraystretch{0.95}

        \caption{Mobile app task complete and partial sequence accuracy on MoTIF with various test splits for evaluating task generalization. The training and test-time input contains the high level goal and low level step by step instructions }
    \begin{tabular}{|l|c|c|c|c|c|c|}
    \hline
   \multirow{5}{*}{Model} & \multicolumn{6}{c|}{MoTIF Test Split}\\
    \cline{2-7}
      & \multicolumn{3}{c|}{App Seen Task Unseen} & \multicolumn{3}{c|}{\multirow{2}{*}{App Unseen Task Seen}}\\
      & \multicolumn{3}{c|}{(Current App)} & \multicolumn{3}{c|}{}\\
    \cline{2-7}
  & \multirow{2}{*}{Action} & \multirow{2}{*}{Ground} & Action + &  \multirow{2}{*}{Action} & \multirow{2}{*}{Ground} & Action +\\
   & & & Ground& & & Ground\\
    \hline
    \textbf{Seq2Seq} & & & & & & \\
    Complete & 75.4 & 31.0 & 31.0 & 70.9 & 25.8 & 25.8\\
    Partial & 92.7 & 46.6 & 46.6 & 91.5 & 41.4 & 41.2 \\ 
    \hline
    \textbf{MOCA} & & & & & & \\
    Complete & 66.5 & 34.3 & 33.1 & 57.9 & 29.5 & 28.1 \\ 
    Partial & 87.8 & 47.7 & 46.2 & 77.8 & 44.7 & 42.7 \\
    \hline
    \end{tabular}
    \label{tab:task_gen}
\end{table}

\end{document}